\documentclass{article}

\PassOptionsToPackage{numbers}{natbib}
\usepackage[final]{neurips_2025}

\usepackage[utf8]{inputenc} %
\usepackage[T1]{fontenc}    %
\usepackage{url}            %
\usepackage{booktabs}       %
\usepackage{amsfonts}       %
\usepackage{nicefrac}       %
\usepackage{microtype}      %
\usepackage{xcolor}         %
\usepackage{graphicx}
\usepackage{amsmath}
\usepackage{multirow}

\usepackage{color,colortbl}
\definecolor{cvprblue}{rgb}{0.21,0.49,0.74}
\usepackage[pagebackref,breaklinks,colorlinks,citecolor=cvprblue]{hyperref}

\newcommand{\figref}[1]{Fig.~\ref{#1}}
\newcommand{\tabref}[1]{Tab.~\ref{#1}}
\newcommand{\eqnref}[1]{Eq.~(\ref{#1})}
\newcommand{\secref}[1]{Sec.~\ref{#1}}

\title{Learning to Generate Human-Human-Object Interactions from Textual Descriptions}

\author{
    Jeonghyeon Na\footnotemark[1], \  Sangwon Baik\footnotemark[1], \ Inhee Lee, \ Junyoung Lee, \ Hanbyul Joo\footnotemark[2] \\
    \\
    Seoul National University
    \\
    \footnotemark[1] \  Equal Contribution \ \ \ \footnotemark[2] \ Corresponding Author \\
    {\tt\small \{prom317,bsw1907,ininin0516,juncong,hbjoo\}@snu.ac.kr}\\
    {\tt\small \href{https://tlb-miss.github.io/hhoi/}{\color{magenta}{https://tlb-miss.github.io/hhoi/}}}\\
    }

\begin{document}

\maketitle

\begin{abstract}
The way humans interact with each other, including interpersonal distances, spatial configuration, and motion, varies significantly across different situations. To enable machines to understand such complex, context-dependent behaviors, it is essential to model multiple people in relation to the surrounding scene context.
In this paper, we present a novel research problem to model the correlations between two people engaged in a shared interaction involving an object. We refer to this formulation as Human-Human-Object Interactions (HHOIs).
To overcome the lack of dedicated datasets for HHOIs, we present a newly captured HHOIs dataset and a method to synthesize HHOI data by leveraging image generative models. As an intermediary, we obtain individual human-object interaction (HOIs) and human-human interaction (HHIs) from the HHOIs, and with these data, we train an text-to-HOI and text-to-HHI model using score-based diffusion model. Finally, we present a unified generative framework that integrates the two individual model, capable of synthesizing complete HHOIs in a single advanced sampling process. Our method extends HHOI generation to multi-human settings, enabling interactions involving more than two individuals.
Experimental results show that our method generates realistic HHOIs conditioned on textual descriptions, outperforming previous approaches that focus only on single-human HOIs. Furthermore, we introduce multi-human motion generation involving objects as an application of our framework.
\end{abstract}

\section{Introduction}
\label{sec:intro}

Human behavior in real-world environments is inherently social and context-dependent. People naturally interact with one another through structured patterns of interpersonal distance, spatial configuration, and motion, which are intuitively understood by humans but vary significantly across different situations. Understanding these nuanced, multi-human interactions is critical for AI systems that aim to interpret, simulate, or engage naturally in human-centric environments.
While Human-Object Interactions (HOIs)~\cite{zhao2022coins,xu2023interdiff,diller2023cghoi,li2024controllable_CHOIS,zhang2024ood,zhang2024hoim3,song2024hoianimator,kim2024beyondcoma} and Human-Human Interactions (HHIs)~\cite{zheng2021deepmulticap,joo2015panoptic,fan2024freemotion,liang2024intergen,shafir2024human,fan2024freemotion,shan2024multiperson,xu2023actformer,tanaka2023role,lim2023mammos,xu2024regennet} have been extensively studied in isolation, modeling interactions involving both multiple people and shared objects remains underexplored. In particular, dyadic interactions, where two individuals engage in a coordinated activity around a common object, are prevalent in everyday life, but have received relatively little attention. Examples include sitting together on a sofa, sharing an umbrella, or standing side by side at a whiteboard for discussion. 
In this paper, we present a novel research problem: modeling the coordinated behavior of two individuals interacting around a shared object. We refer to this formulation as Human-Human-Object Interactions (HHOIs). A core challenge in studying HHOIs is the absence of dedicated datasets. Existing HOI datasets~\cite{3DPW,guzov2021human,zheng2022gimo,pons2023interaction,zhang2023neuraldome} primarily feature single-human-object interactions, while HHI datasets~\cite{kundu2020cross,fieraru2020three,guo2022multi,liang2024intergen,xu2024interx,ghosh2024remos,muller2024generative_proxemics,shafir2024human} typically lack object context, often limited to dyadic conversational scenarios in standing poses.

To address the lack of diverse HHOI data, we introduce a newly collected 3D dataset captured using a multi-camera system, specifically designed to support the training and evaluation of HHOI models.
In addition, we present a synthetic HHOI dataset generation pipeline that leverages pretrained image diffusion models~\cite{rombach2022high} to complement real-world data, particularly for scenarios that are challenging to capture in studio environments.
These diverse data sources are unified through a score-based diffusion model, enabling realistic generation of dyadic human-object interactions.
Our model is conditioned on textual descriptions, enabling semantically grounded generation of HHOIs. 
Importantly, we further demonstrate that our framework can be extended beyond dyadic interactions to accommodate multi-human interactions, offering a scalable solution for modeling increasingly complex social behaviors. Experimental results show that our method produces more realistic and coherent interactions compared to existing baselines that model only single-human HOIs. As a potential application of our model, we present multi-human motion generation via Diffusion-Noise-Optimization~\cite{karunratanakul2023dno}, with our output HHOI as constraint.

Our contributions are summarized as follows:  
(1) Curated datasets for HHOI, along with methodologies for constructing them; 
(2) Score-based HHOI model that jointly captures both individuals’ interactions with a shared object and their interaction with each other, conditioned on textual description; (3) Extension to multi-human HHOIs, enabling generation of interactions beyond dyadic settings; (4) Application to object interaction-aware multi-human motion generation.

\begin{figure}
    \centering
    \includegraphics[width=\linewidth, trim={0cm 0cm 0cm 0cm},clip]{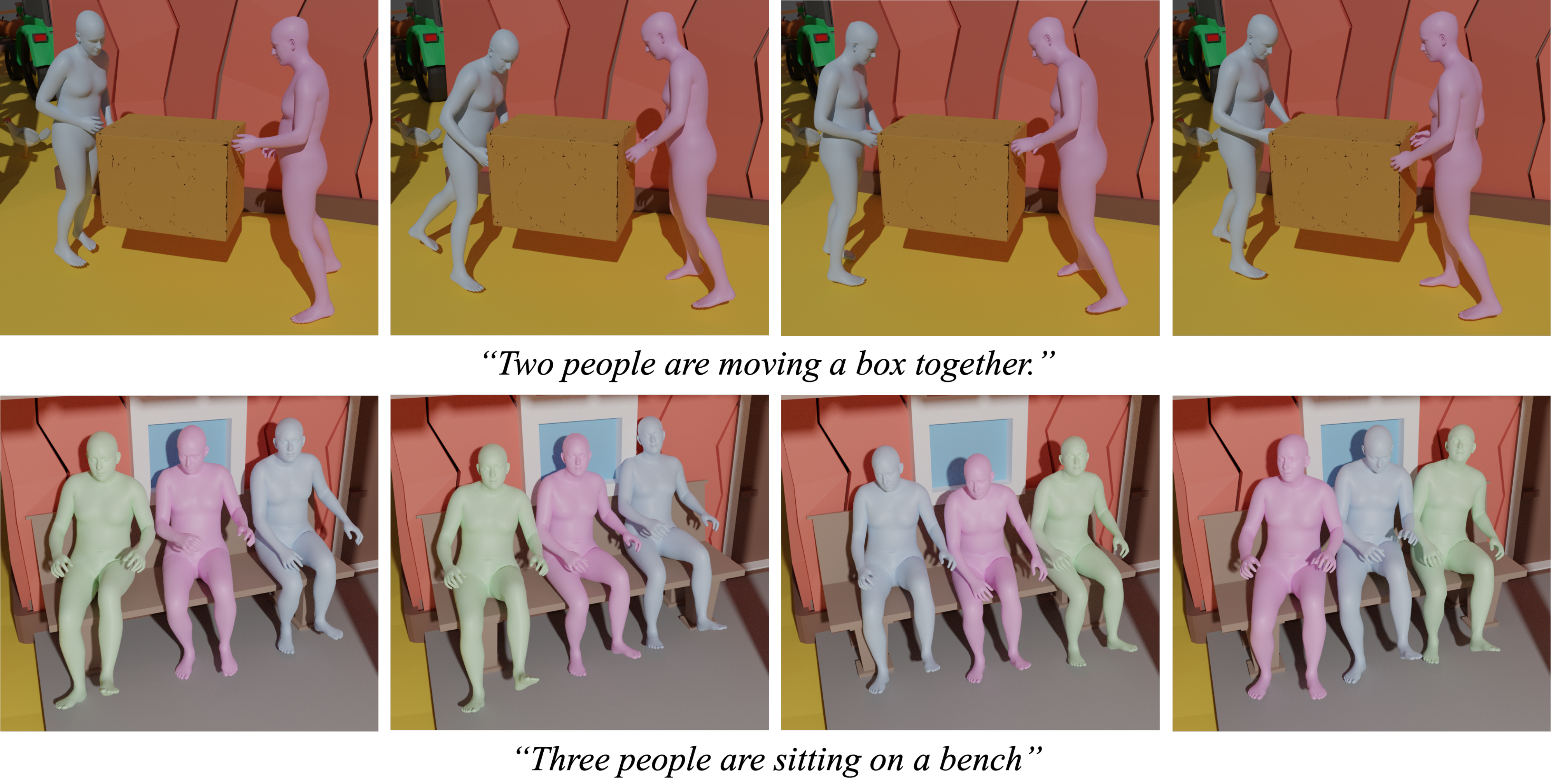}
    \vspace{-0.4cm}
    \caption{Results of our HHOI generation given object instances and text-prompt descriptions. Multiple humans in action are generated by jointly enforcing scene-level consistency across human-object interactions (HOIs) and human-human interactions (HHIs).}
    \label{fig:teaser}
\end{figure}

\section{Related Work}
\label{sec:relwork}

\noindent\textbf{Human-Object Interaction}
Human-object interaction (HOI) aims to understand how humans interact with objects in the environment. This line of research is crucial for enabling machines to interpret and mimic human behaviors, thereby supporting the development of embodied AI agents and realistic digital human modeling with natural object manipulation.
There have been considerable efforts to collect and scale up 3D HOI dataset, aiming to pursue a data-driven approach in this direction. Early work tries to capture the 3D HOI scenes in a controlled setup
using marker~\cite{jiang2024scaling}, IMU~\cite{3DPW,guzov2021human,zheng2022gimo,pons2023interaction}, multi-view capture system~\cite{RICH,huang2022intercap,zhang2023neuraldome,bhatnagar2022behave}, or hybridizing them~\cite{fan2023arctic,kim2024parahome,zhang2024force,zhang2024hoim3,liu2024core4d}.
In addition to real-world capture, synthetic datasets have also been introduced using game engines~\cite{cao2020long} and physics simulation~\cite{black2023bedlam}. 
More recently, automated pipelines have emerged to generate 3D HOI scenes from pre-trained 2D image models, significantly reducing capture effort and expanding scenario diversity~\cite{kim2024beyondcoma,CHORUS,genzi,david}.
Various models~\cite{xu2023interdiff, diller2023cghoi, li2024controllable_CHOIS, zhang2024ood, song2024hoianimator, xu2024interdreamer, kulkarni2024nifty, yi2024tesmo, jiang2024autonomous} have been proposed to learn from the presented datasets, including models manipulating articulated objects ~\cite{fan2023arctic,kim2024parahome}, and multiple objects at once ~\cite{zhang2024hoim3,kim2024parahome}. However, the majority of existing methods are limited to single-human interaction scenarios, while scenarios involving multiple humans with objects remain largely unexplored.
Core4D~\cite{liu2024core4d} takes a step toward this direction by collecting HHOI data for collaborative tasks involving two people, but its scale and diversity are still limited.

\noindent\textbf{Human-Human Interaction} Modeling the interaction between individuals is essential to capture cooperative behaviors and social dynamics~\cite{joo2019towards}, which is crucial for developing embodied AI agents capable of natural human interaction. 
Previous works have primarily focused on dyadic human-human interactions, aiming either to reconstruct plausible interactions from images~\cite{fieraru2020three, muller2024generative_proxemics}, or to generate natural motion~\cite{kundu2020cross, liang2024intergen, xu2024interx, ghosh2024remos, shafir2024human}. Recently, several methods have been proposed to extend the interaction modeling to more than two individuals~\cite{zheng2021deepmulticap, joo2015panoptic, fan2024freemotion, shan2024multiperson}. 
Many of these approaches are conditioned by contextual signals such as text~\cite{liang2024intergen,shafir2024human,fan2024freemotion,shan2024multiperson}, music~\cite{siyao2024duolando,li2024interdance}, or predefined action-reaction roles~\cite{xu2023actformer,tanaka2023role,lim2023mammos,xu2024regennet}, yet they often struggle to capture interactions that are tightly coupled with a specific target object. MAMMOS~\cite{lim2023mammos} explores scene-conditioned multi-human motion generation, yet remains limited in scope and lacks complex interactions.

\noindent\textbf{Score-based Generative Models.} 
Score-based generative models~\cite{song2019generative,song2020improved} estimate gradients of the data distribution for generative modeling, by introducing noise conditional score networks to learn score functions at multiple noise levels. Later work~\cite{song2020score} generalized this approach using stochastic differential equations, providing a continuous-time formulation. This framework has proven highly effective for various generation tasks such as object rearrangement~\cite{wu2022targf}, object pose estimation~\cite{genpose}, scene-graph generation~\cite{energyscene} and human pose estimation~\cite{ci2022gfpose}. Recent extensions apply score-based models to interactive settings, such as human-object interaction~\cite{david}, and object-object interaction~\cite{oor}.

\begin{figure}[t]
    \centering
    \includegraphics[width=\linewidth, trim={0 0 0 0},clip]{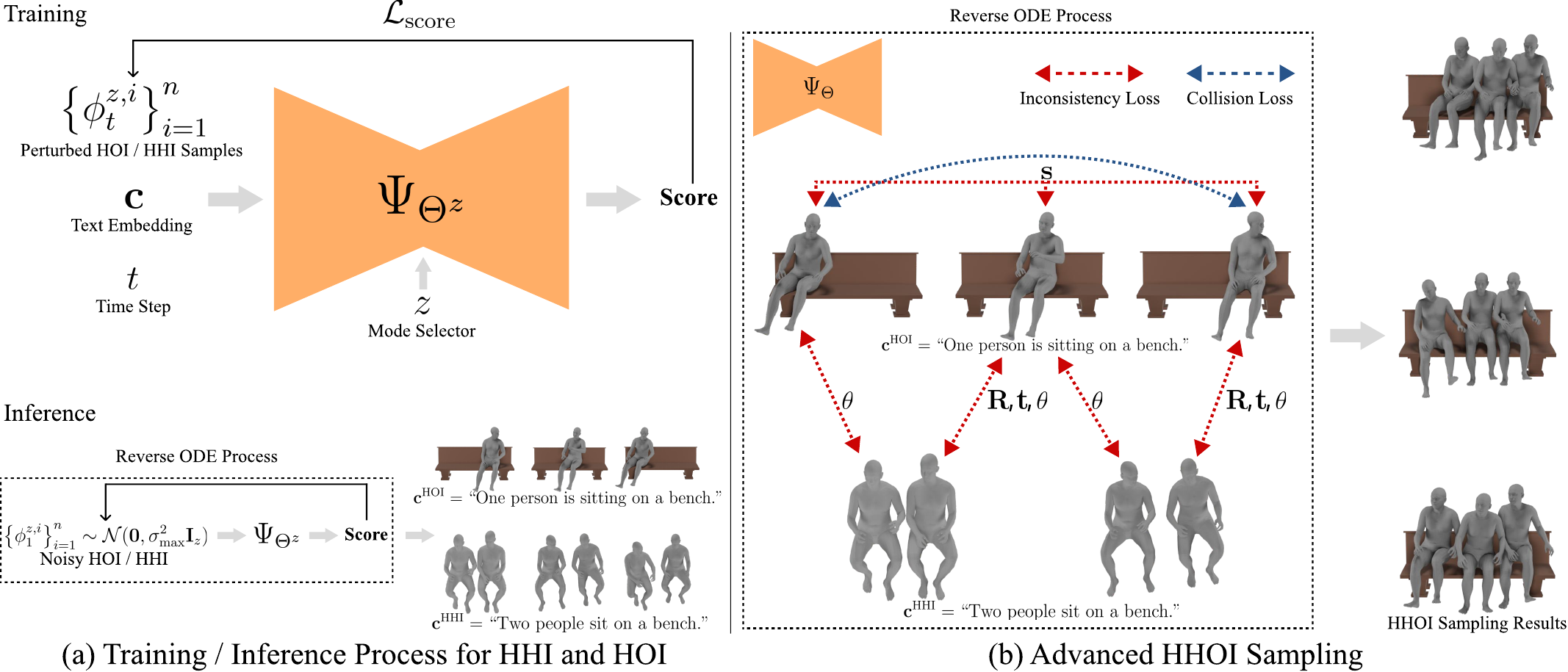}
    \caption{\textbf{Method Overview.} (a) The training and inference process of the HOI/HHI part. (b) The advanced HHOI sampling process by introducing inconsistency loss and collision loss.}
    \label{fig:model_overview}
    \vspace{-15pt}
\end{figure}

\vspace{-12pt}
\section{Method}
\label{sec:method}

Our HHOI model is structured as a combination of HOI model and HHI model.  
We first introduce how each component—HOI model and HHI model—is independently represented (\secref{sec:hhoi_formulation}).
Specifically, we model HOI and HHI using score-based diffusion models~\cite{song2019generative,song2020score}. We then describe the training and inference procedures for each component (\secref{sec:hhoi_diffusion}).
Finally, we present a guided sampling method that integrates each component with inconsistency and collision constraints during the inference process to generate coherent and plausible HHOI configurations (\secref{sec:hhoi_infer}).

\subsection{HHOI Formulation}
\label{sec:hhoi_formulation}

\noindent \textbf{Modeling Human-Object Interaction (HOI).}
For HOI, we adopt an object-centric coordinate frame in which the object instance mesh $\mathcal{M}$ is centered at the origin. Our goal is to model the distribution of plausible human spatial and postural configurations relative to the object, covering a variety of HOI scenarios.
Formally, we define HOI as the rotation $\mathbf{R}_\mathcal{H} \in \mathbb{R}^6$, translation $\mathbf{t}_\mathcal{H} \in \mathbb{R}^3$, scale $\mathbf{s}_\mathcal{H} \in \mathbb{R}_+$, and body pose embedding $\theta_\mathcal{H} \in \mathbb{R}^{H}$ of a human $\mathcal{H}$. These are conditioned on an object mesh $\mathcal{M}$ and a textual description $\mathbf{c}$ describing the HOI.
We denote the HOI distribution as $p_\mathbf{c}^{\mathcal{M}}$, and a corresponding HOI sample as $\phi^{\text{HOI}}$:
\begin{gather}
    \mathbf{\phi}^{\text{HOI}} \sim p_\mathbf{c}^{\mathcal{M}}, \; \; \mathbf{\phi}^{\text{HOI}} = 
    (\mathbf{R}_\mathcal{H}, \mathbf{t}_\mathcal{H}, \mathbf{s}_\mathcal{H}, \theta_\mathcal{H}).
    \label{eq:hoi_formulation}
\end{gather}
We use SMPL-X~\cite{smplx} for the human model and $6$D representation~\cite{6d_repr} for $\mathbf{R}_\mathcal{H}$. Given an HOI sample $\phi^{\text{HOI}}$, human mesh $\mathcal{H}$ can be obtained as follows:
\begin{equation}\label{eqn:hoi_to_human}
    \begin{gathered}
        \mathcal{H} = \mathbf{s}_\mathcal{H} \cdot \mathbf{R}_\mathcal{H} \mathcal{H}^{\text{cano}} + \mathbf{t}_\mathcal{H}, \\
        \mathcal{H}^{\text{cano}} = \mathbf{smplx}(\mathbf{dec}(\theta_\mathcal{H})),
    \end{gathered}
\end{equation}
where $\mathbf{dec}(\cdot)$ is the body pose decoder (detailed below) that maps the body pose embedding $\theta_\mathcal{H}$ to SMPL-X pose $\theta\in\mathbb{R}^{21\times6}$. The function $\mathbf{smplx}(\theta)$ returns the SMPL-X mesh for pose $\theta$ in its canonical frame. We use by $\mathbf{R}_\mathcal{H}$ both the $6$D representation and the corresponding SO(3) rotation, since there is a straightforward one-to-one mapping between them. 

\noindent \textbf{Modeling Human-Human Interaction (HHI).}
Given a text prompt $\mathbf{c}$ that describes the interaction between two humans, denoted $\mathcal{H}_1$ and $\mathcal{H}_2$, we model the HHI using their body poses along with the relative rotation and translation of $\mathcal{H}_2$ with respect to $\mathcal{H}_1$. We assume both humans share the same scale. Specifically, we define the HHI distribution as $p_\mathbf{c}^{\mathcal{H} \rightarrow \mathcal{H}}$ and the HHI sample as $\phi^{\text{HHI}}$:
\begin{gather}
    \mathbf{\phi}^{\text{HHI}} \sim p_\mathbf{c}^{\mathcal{H} \rightarrow \mathcal{H}}, \; \; \mathbf{\phi}^{\text{HHI}} = 
    (\theta_{\mathcal{H}_1}, \mathbf{R}_{\mathcal{H}_2 \rightarrow \mathcal{H}_1}, \mathbf{t}_{\mathcal{H}_2 \rightarrow \mathcal{H}_1}, \theta_{\mathcal{H}_2}).
    \label{eq:hhi_formulation}
\end{gather}
The human $\mathcal{H}_2$ can be reconstructed from the HOI of $\mathcal{H}_1$ and the HHI:
\begin{equation}\label{eqn:hhi_to_human}
    \begin{gathered}
        \mathcal{H}_2 = \mathbf{s}_{\mathcal{H}_1} \cdot \mathbf{R}_{\mathcal{H}_1} \mathbf{R}_{\mathcal{H}_2 \rightarrow \mathcal{H}_1} \mathcal{H}^{\text{cano}}_2 + \mathbf{s}_{\mathcal{H}_1} \cdot \mathbf{R}_{\mathcal{H}_1} \mathbf{t}_{\mathcal{H}_2 \rightarrow \mathcal{H}_1} + \mathbf{t}_{\mathcal{H}_1}, \\
        \mathcal{H}^{\text{cano}}_2 = \mathbf{smplx}(\mathbf{dec}(\theta_{\mathcal{H}_2})),
    \end{gathered}
\end{equation}

\textbf{Body Pose Embedding.} We find that modeling the human pose distribution using a low-dimensional embedding $\theta_\mathcal{H}$ is more effective than using the full $126$D (=$21\times6$, $6$D representation for each joint rotation) body pose $\theta$ directly. Therefore, we train a body pose encoder and decoder to obtain an embedding vector of the body pose, enabling us to model HHOI in the latent space of body poses. To this end, we process $922$K human body pose data from ~\cite{RICH, H36M, 3DPW}, and train a body pose encoder and decoder, each implemented as a $4$-layer MLP. In experiments, we embed $126$D human body poses in a $10$D space, that is, $H=10$, which results in $\phi^{\text{HOI}} \in \mathbb{R}^{20}$ and $\phi^{\text{HHI}} \in \mathbb{R}^{29}$.

\subsection{Score-based HHOI Diffusion Model}
\label{sec:hhoi_diffusion}

We model HHOI using score-based diffusion, similar to how poses and scales of objects are modeled in~\cite{oor, genpose}. Let us denote our HHOI diffusion model as $\Psi_\Theta$, parameterized by $\Theta$. Then, $\Psi_\Theta$ represents the noised score function of the HOI, HHI distribution at time step $t$:
\begin{equation}\label{eqn:hhoi_scorenet}
   \Psi_{\Theta^z}(\phi^{z}_{t}, t | \mathbf{c}, z) = \begin{cases}
        \nabla_{\phi^{\text{HOI}}_{t}} \log p_\mathbf{c}^\mathcal{M}(\phi^{\text{HOI}}_{t}), & z = \text{HOI} \\
        \nabla_{\phi^{\text{HHI}}_{t}} \log p_\mathbf{c}^{\mathcal{H} \rightarrow \mathcal{H}}(\phi^{\text{HHI}}_{t}), & z= \text{HHI}
    \end{cases},
\end{equation}
where $\phi_{t}^{(\cdot)}$ is a noised HOI or HHI sample at time step $t$, $z \in \left\{\text{HOI}, \text{HHI}\right\}$ represents mode selector, and $\Theta^{\text{HOI}} \cup \Theta^{\text{HHI}} = \Theta$, $\Theta^{\text{HOI}} \cap \Theta^{\text{HHI}} = \emptyset$. For simplicity, we do not use a mesh instance $\mathcal{M}$ as input when modeling HOI; rather, we assume a fixed $\mathcal{M}$ is provided for each scenario.

\textbf{Training.} As HHOI is formulated by decomposing it into HOI and HHI in Sec.~\ref{sec:hhoi_formulation}, we also train the score-based diffusion in a decoupled manner. Each mode is trained with the following objective function presented in Denoising Score Matching(DSM)~\cite{DSM}:
\begin{equation}\label{eqn:score_loss}
    \begin{gathered}
       \mathcal{L}_{\text{score}}(\Theta^z) = \mathbb{E}_{t \sim \mathcal{U}(\epsilon, 1)}\left[
            \lambda_{t}\mathbb{E}_{
                \phi^{z}, \phi_{t}^{z}
            }\left[ \left\| \Psi_{\Theta^{z}}(\phi_{t}^z, t | \mathbf{c}, z) - \frac{\phi^z - \phi_{t}^z}{\sigma(t)^2} \right\|_2^2 \right]
        \right],
    \end{gathered}
\end{equation}
where $\epsilon$ is minimal noise level, $\phi_{t}^z \sim \mathcal{N}(\phi^z, \sigma^2(t) \mathbf{I}_z)$, $\sigma^2(t) = \sigma_{\text{min}}(\frac{\sigma_{\text{max}}}{\sigma_{\text{min}}})^t$ is a variance factor, and $\lambda_{t}$ is a regularization term.

The training process is shown in Fig.~\ref{fig:model_overview}(a). First, we generate perturbed samples $\phi_{t}^z$ as defined. We then compute the target score $\frac{\phi^z - \phi_{t}^z}{\sigma(t)^2}$ and the estimated score from the HHOI diffusion model to evaluate the score loss in \eqnref{eqn:score_loss}.
We use the CLIP text encoder~\cite{radford2021learning_clip} to obtain a text embedding from a text prompt. For simplicity, we use $\mathbf{c}$ to denote both the text prompt and its corresponding embedding. Additionally, we adopt the text prompt augmentation strategy from~\cite{oor}.

\textbf{Inference.} To sample HOI and HHI instances independently, we solve the following Probability Flow(PF) ODE~\cite{song2020score} in reverse time, i.e., from $t=1$ to $t=\epsilon$:
\begin{equation}\label{eqn:reverse_ode}
    \begin{gathered}
       \phi_{1}^z \sim \mathcal{N}(\mathbf{0}, \sigma_{\text{max}}^2\mathbf{I}_z), \\
       \frac{d\phi_{t}^z}{dt} = -\sigma(t)\dot\sigma(t)\Psi_{\Theta^{z}}(\phi_{t}^z, t | \mathbf{c}, z).
    \end{gathered}
\end{equation}
We solve the ODE using external library~\cite{torchdiffeq}, which is fully supported to run on the GPU. Fig.~\ref{fig:model_overview}(a) shows this process. However, independently sampling HOI and HHI leads to incoherent posture and spatial configuration in each sampling output, and does not give us the desired HHOI.
To address this issue, we propose an advanced guided sampling technique to obtain HHOI in Sec.~\ref{sec:hhoi_infer}.

\subsection{Advanced Guided Sampling for HHOI}
\label{sec:hhoi_infer}

We adopt the PF ODE augmentation paradigm~\cite{oor,david}, introducing additional terms that enforce the appropriate combination of HOI and HHI samples during HHOI sampling. Specifically, we incorporate an inconsistency loss to enforce consistency across samples and a collision loss to prevent human collisions. Consider the HHOI involving $N$ humans(i.e., $N$ HOIs), along with $M$ HHIs:
\begin{equation}\label{eqn:hhoi_example}
    \begin{gathered}
       \mathbf{\phi}^{\text{HOI}, \mathcal{H}_i} \sim p_{\mathbf{c}^{\text{HOI}}}^{\mathcal{M}}, \; \; \mathbf{\phi}^{\text{HOI}, \mathcal{H}_i} = 
        (\mathbf{R}_{\mathcal{H}_i}, \mathbf{t}_{\mathcal{H}_i}, \mathbf{s}_{\mathcal{H}_i}, \theta_{\mathcal{H}_i}), \\
        \mathbf{\phi}^{\text{HHI}, \mathcal{H}_k \rightarrow \mathcal{H}_j} \sim p_{\mathbf{c}^{\text{HHI}}}^{\mathcal{H} \rightarrow \mathcal{H}}, \; \; \mathbf{\phi}^{\text{HHI}, \mathcal{H}_k \rightarrow \mathcal{H}_j} = 
        (\theta_{\mathcal{H}_j}, \mathbf{R}_{\mathcal{H}_k \rightarrow \mathcal{H}_j}, \mathbf{t}_{\mathcal{H}_k \rightarrow \mathcal{H}_j}, \theta_{\mathcal{H}_k}),
    \end{gathered}
\end{equation}
where $i = 1, \dots, N$, and $|\left\{(j, k) \; | \; 1 \leq j, k \leq N, \; j \neq k\right\}| = M \leq \frac{N(N-1)}{2}$. Note that the number of possible HHIs is at most $_NC_2 =\frac{N(N-1)}{2}$. 
The graph formed by connecting the $M$ human pairs should be a directed acyclic graph (DAG).
For example, given three HHIs $\left\{\mathcal{H}_2 \rightarrow \mathcal{H}_1, \mathcal{H}_3 \rightarrow \mathcal{H}_2, \mathcal{H}_1 \rightarrow \mathcal{H}_3\right\}$, this is not a valid HHI set because it has a circular dependency. 

\textbf{Inconsistency Loss.}
To generate a unified HHOI sample from inconsistency HOI/HHI samples set, we introduce inconsistency loss $\mathcal{L}_{\text{inc}}$ enforcing coherence between the human representations derived by each sample. In a nutshell, $\mathcal{L}_{\text{inc}}$ is the role of enforcing consistency between humans obtained from \eqnref{eqn:hoi_to_human} and humans obtained from \eqnref{eqn:hhi_to_human}. Specifically, it penalizes discrepancies in scale, body pose, rotation, and translation of HOI and HHI samples at time step $t$ during sampling, as follows:
\begin{equation}\label{eq:inc_loss_overview}
    \mathcal{L}_{\text{inc}}(\Phi_t) = 
     \mathcal{L}_{var,s}(\mathbf{s}) +  \mathcal{L}_{var,\theta}(\theta) +  \mathcal{L}_{var,R}(\mathbf{R}) +  \mathcal{L}_{var,t}(\mathbf{t}),
\end{equation}
where $\Phi_t$ denotes union of $N$ HOI samples $\{\phi_t^{{\text{HOI}}, \mathcal{H}_i}\}$ and $M$ HHI samples $\left\{\mathbf{\phi}^{\text{HHI}, \mathcal{H}_k \rightarrow \mathcal{H}_j}_t\right\}$ at time step $t$.
Each term in \eqnref{eq:inc_loss_overview} minimizes the variance of each component, thereby enhancing overall consistency as detailed below.

The scale variance loss, $\mathcal{L}_{var,s}(\mathbf{s})$ penalizes deviations in human scale across HOI samples since the HHI model assumes equal scales for paired humans:
\begin{equation}
\mathcal{L}_{var,s}(\mathbf{s})=N\cdot\mathrm{Var}(\{\mathbf{s}_{\mathcal{H}_i}\}_{i=1}^N).
\end{equation}
The body pose variance loss $\mathcal{L}_{var,\theta}(\theta)$ enforces consistency in body poses for the same human:
\begin{equation}
\mathcal{L}_{var,\theta}(\theta) = \sum_{i=1}^N N_i \cdot \mathrm{Var}(\{\theta_{\mathcal{H}_i, n}\}_{n=1}^{N_i}),
\end{equation}
where $N_i$ denotes total occurrences of human $\mathcal{H}_i$ across HOI and HHI samples. More precisely, $N_i - 1$ is the number of HHIs where $\mathcal{H}_i$ appears.
The rotation and translation variance losses, $\mathcal{L}_{var,R}(\mathbf{R})$ and $\mathcal{L}_{var,t}(\mathbf{t})$, ensure consistency among human rotation and translation across interactions:
\begin{equation}
\mathcal{L}_{var,R}(\mathbf{R}) = \sum_{i=0}^N N'_i \cdot \mathrm{Var}(\{\mathbf{R}_{\mathcal{H}_i}\}\cup\{\mathbf{R}_{\mathcal{H}_{jn}}\mathbf{R}_{\mathcal{H}_i \rightarrow \mathcal{H}_{jn}}\}_{n=1}^{N'_i}),
\end{equation}
\begin{equation}
\mathcal{L}_{var,t}(\mathbf{t}) = \sum_{i=0}^N N'_i \cdot \mathrm{Var}(\left\{\mathbf{t}_{\mathcal{H}_i}\right\} \cup \left\{\mathbf{s}_{\mathcal{H}_{jn}} \cdot \mathbf{R}_{\mathcal{H}_{jn}}\mathbf{t}_{\mathcal{H}_i \rightarrow \mathcal{H}_{jn}} + \mathbf{t}_{\mathcal{H}_{jn}}\right\}_{n=1}^{N'_i}),
\end{equation}
where $N'_i$ represents the number of HHIs where human $\mathcal{H}_i$ appears as a target(i.e., in pairs of the form $\mathcal{H}_i \rightarrow \mathcal{H}_j$).
Note that $\mathrm{Var}(\left\{x_1, \dots, x_n\right\}) = \frac{1}{n} \sum_{i=1}^n (x_i - \bar{x})^2$.

\textbf{Collision Loss.} To ensure plausible interactions among multiple humans, it is essential to avoid physically implausible scenarios such as unintended body intersections. In particular, for human pairs that are not explicitly connected by a HHI, we assume an “no collision” constraint—that is, the absence of collision serves as their implicit HHI. For instance, if the HHI set is $\{\mathcal{H}_2 \rightarrow \mathcal{H}_1, \mathcal{H}_3 \rightarrow \mathcal{H}_2\}$, then we enforce that $\mathcal{H}_1$ and $\mathcal{H}_3$ do not collide.

However, computing accurate inter-human collisions using full SMPL-X meshes during ODE-based sampling is computationally expensive. To address this, we approximate collisions between two humans in following steps:
(1) Compute joint positions via forward kinematics over body joints using each human pose;
(2) Construct a 24-capsule proxy from these joints with predefined radius factors to approximate each human as a capsule-based model;
(3) Compute the sum of overlaps over all capsule pairs for two humans as the collision loss.
Each overlap is computed simply as the sum of the two radii minus the distance between the capsules’ axis segments.
Accordingly, the collision loss $\mathcal{L}_{\text{col}}(\Phi_t)$ is computed as
\begin{equation}\label{eq:col_loss_v2}
    \mathcal{L}_{\text{col}}(\Phi_t) = \displaystyle\sum_{(\mathcal{H}_i, \mathcal{H}_j) \in \Phi_{\text{nap}}} \frac{1}{24^2} \sum_{c_i=1}^{24} \sum_{c_j=1}^{24} \max(0, r_{c_i}^{\mathcal{H}_i} + r_{c_j}^{\mathcal{H}_j} - d_{c_i, c_j}^{\mathcal{H}_i, \mathcal{H}_j}),
\end{equation}
where $\Phi_{\text{nap}}$ is the set of non-adjacent human pairs, $r_{c_i}^{\mathcal{H}_i}$ is the radius of the $c_i$-th capsule of human $\mathcal{H}_i$, and $d_{c_i, c_j}^{\mathcal{H}_i, \mathcal{H}_j}$ is the distance between the axis segment of the $c_i$-th capsule of human $\mathcal{H}_i$ and the axis segment of the $c_j$-th capsule of human $\mathcal{H}_j$. Note that in \eqnref{eq:col_loss_v2}, if the sum of the two capsule radii is less than the distance between their axis segments, there is no collision and the value is clipped to zero. See Sec.~\ref{sec:capsule_appendix} for a process to approximate a human with a 24-capsule proxy.

\textbf{Guided HHOI Sampling.} Using \eqnref{eq:inc_loss_overview} and \eqnref{eq:col_loss_v2}, we augment the PF ODE in \eqnref{eqn:reverse_ode} as follows to sample HHOI:
\begin{equation}\label{eqn:reverse_ode_full}
    \begin{gathered}
       \phi_{1}^{z, i} \sim \mathcal{N}(\mathbf{0}, \sigma_{\text{max}}^2\mathbf{I}_z), \\
       \frac{d\phi_{t}^{z, i}}{dt} = -\sigma(t)\dot\sigma(t)\Psi_{\Theta^{z}}(\phi_{t}^{z, i}, t | \mathbf{c}^z, z) + 
       \lambda_1\nabla_ {\phi_{t}^{z, i}}\mathcal{L}_{\text{inc}}(\Phi_t) +
       \lambda_2\nabla_{\phi_{t}^{z, i}}\mathcal{L}_{\text{col}}(\Phi_t),
    \end{gathered}
\end{equation}
where $\phi^{z, i}_t$ is $i_\text{th}$ HOI($z=\text{HOI}$) sample or HHI($z=\text{HHI}$) sample at time step $t$, $\lambda_1$ and $\lambda_2$ are weight terms. In ~\figref{fig:model_overview}(b), we provide an intuitive overview of our advanced sampling process through a concrete HHOI example in which three humans sit on a bench.

\section{Datasets}
\label{sec:dataset}

In this section, we present the motivation and data collection procedure for our dataset. 
To align with our model architecture, the distribution of body poses in the HOI dataset must be consistent with that of the HHI dataset.
For instance, HOI dataset with action ``sitting on a bench'', may consist only of body pose with head pose fixed to the front, assuming only one human is in the scene, while the HHI dataset may contain a broader range of head orientations more representative of real-world variability. In this case, our model will be unable to generate these natural variations during the HHOI synthesis.

The most straightforward solution is to collect data from scenes that include both multiple humans and interacting objects. While several individual datasets that contain either multi-human poses or human-object interactions have been released, there are a limited number of datasets that capture both multi-human poses and human-object interactions.  CORE4D~\cite{liu2024core4d} is one of them, providing high-quality motion capture data of two individuals interacting with a single object. It includes object pose annotations and human SMPL-X parameters for six object categories. 
However, the range of interaction types in CORE4D is relatively narrow, focusing primarily on collaborative actions such as passing an object or moving it together. In contrast, real-world interactions are more diverse. Notably, there are scenarios where no direct contact occurs between humans and objects, yet the mere presence of the object influences spatial relationships (proxemics) and interaction dynamics between individuals. Such implicit object effects are underrepresented in existing datasets. 

To address the lack of diverse human-human-object interactions, we collect dyadic human and object poses using a multi-view camera system. Specifically, we record two people interacting with various objects from 36 synchronized RGB cameras.
We apply an off-the-shelf 2D pose detector~\cite{dwpose} and a 3D optimization method~\cite{mvsmplifyx} to recover the SMPL-X parameters for both individuals in each camera frame, followed by manual post-filtering. 
The collected HHOI dataset is divided into HOI and HHI subsets, which are used to train the respective score-based models. The same data processing and training procedure is applied to the CORE4D dataset.
Our dataset consists of 5,078 frames spanning 11 object categories, and is expressed in metric units. See Sec.~\ref{sec:data_collection} for more details.

While the aforementioned datasets provide high-quality HHOI data, their content is limited to scenarios that can be captured in controlled studio environments. As a result, interactions involving large objects or outdoor scenarios, such as “two people riding a motorcycle together”, are difficult to capture with our multi-view camera systems. To address this limitation, we leverage the knowledge of 2D image diffusion model on HOI and HHI. Recent models such as Flux~\cite{flux2024blackforest} can generate realistic human-object interaction images from textual prompts while preserving crucial social cues, including joint body poses and interpersonal distances that reflect real-world dynamics.

To generate HOI data, we utilize ComA~\cite{kim2024beyondcoma}, which takes as input an object instance and a textual description of an HOI scenario, and outputs corresponding 3D human-object interactions represented via SMPL-X parameters. For HHI data, we generate images of dyadic human interactions involving an object using textual prompts and the Flux diffusion model. Subsequently, we apply a Human Mesh Recovery method~\cite{multihmr} to reconstruct 3D human meshes from the generated images.

\section{Experiments}
\label{sec:experiments}

\subsection{Baselines and Metrics}
\label{sec:baseline_metric}
\textbf{Baselines.} Previous research on human-object interaction has primarily focused on generating a single human pose with respect to a specific object. As there exists no comparable method for generating multiple humans interacting with an object as in our setting, we develop two approaches for this novel task: (1) by extending GenZI~\cite{genzi}, and (2) by lifting separately generated humans using a depth optimization strategy, referred to as Depth Opt.

First, we modify GenZI, which originally renders a target object in multi-view and inpaints a single human based on a text prompt. We revise the prompt and inpainting process to synthesize images with multiple people, and follow GenZI’s pipeline by detecting 2D body poses and optimizing SMPL-X parameters using multi-view joint positions. For Depth Opt., we inpaint a single-view object rendering using a diffusion-based model~\cite{rombach2022high}, extract 3D human meshes via Human Mesh Recovery~\cite{multihmr}, and align them to the object using depth optimization with Depth-Pro~\cite{ml-depth-pro}. See Sec.~\ref{sec:exp_detail} for further details.

\begin{figure}[!t]
    \centering
    \includegraphics[width=\linewidth, trim={0cm 0.0cm 0cm 0cm},clip]{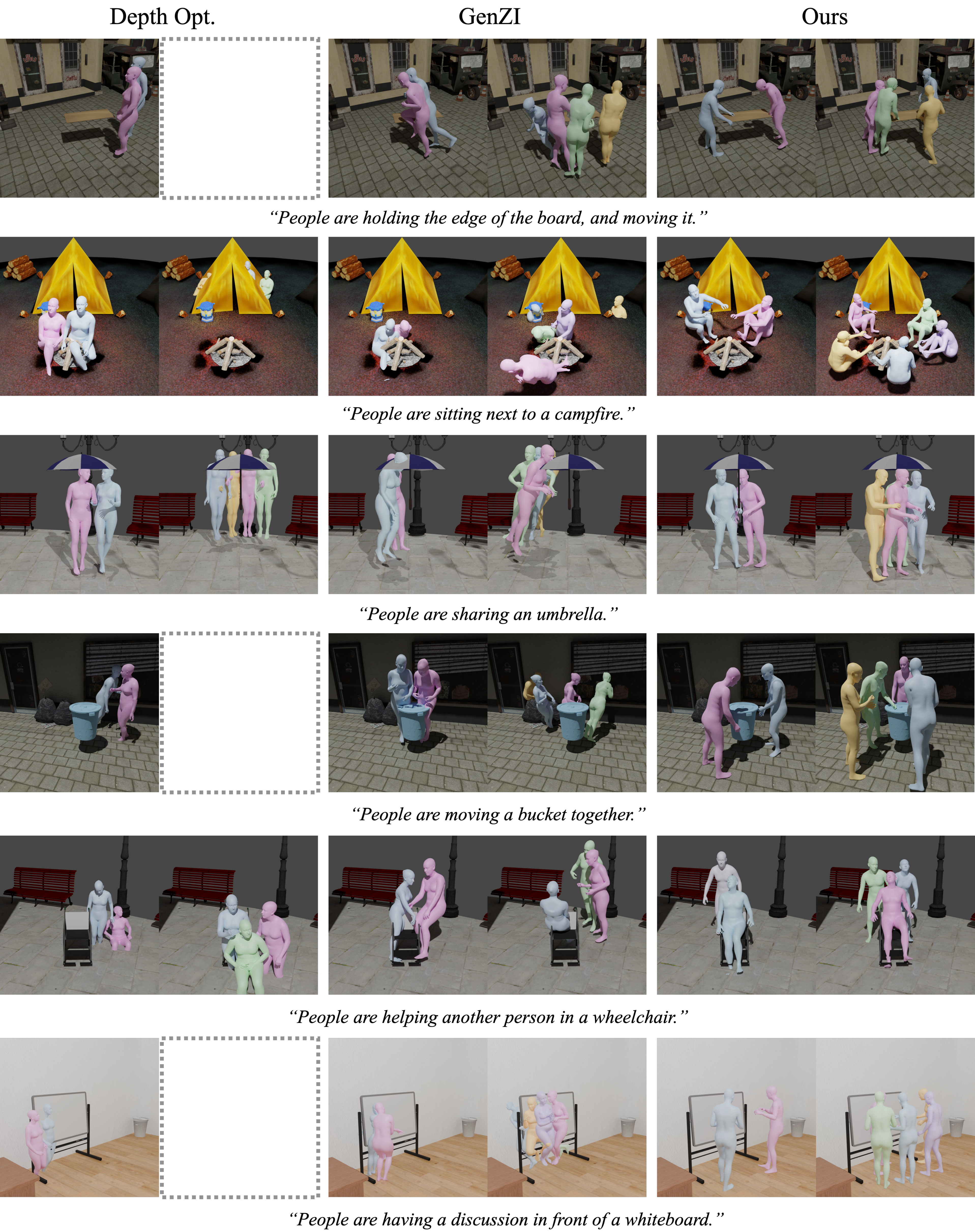}
    \caption{HHOI generation result of dyadic, and multiple humans in action with our model and baselines. In multiple HHOI, number of humans ranges from 3 to 5. Empty result represents cases where generation failed in 10 trials. Our model can generate complex HHOIs with varying number of humans in the scene, while preserving the natural social cues.}
    \label{fig:qual_two_person}
    \vspace{-10pt}
\end{figure}

\textbf{Metrics.} To evaluate how realistically our model generates body poses and interpersonal distances, we compare the distributions of generated results against the test set in CORE4D and our collected dataset. For body pose, we compute the Fréchet Distance (FD) between the embedded pose distributions which are obtained via our body pose encoder. For interpersonal distance, we calculate FD on the per-human global translation differences. 

We additionally employ CLIP-score~\cite{radford2021learning_clip} to assess semantic alignment between the generated outputs and the input text prompts. The HHOI outputs are rendered from multi-view, and CLIP-score is computed by averaging image-text cosine similarities across these views. For cases involving more than three humans, we report the success rate to evaluate the robustness of our model in multi-human settings. We acccount generation success as fully generating desired number of people in 3D. 

Physical plausibility of the generated HHOI is also an important factor, so we evaluate it using two metrics—penetration ratio and contact distance. Penetration ration measures the ratio of mesh vertices that lie inside another mesh, against the total mesh vertices. We count human vertices inside another human or the object and report human-human and human-object separately. Contact distance is measured only for categories requiring contact, split into hand-contact (board, box, bucket, chair, desk) and hip-contact (bench). For each, we randomly sample 10 vertices from the relevant body parts and compute their mean distance to the object mesh.

Finally, we conduct a user study to further validate the realism and text coherence of our method. Participants compare outputs from our model and baselines and select the most realistic and faithful to the prompt.

\begin{table}
  \caption{Quantitative comparison on text-guided dyadic HHOI generation. Our model shows robust performance compared to baseline models in all the metrics.}
  \label{two_people_table}
  \centering
  \begin{tabular}{lcccc}
    \toprule
    Method     & Body Pose FD $\downarrow$ & Distance FD $\downarrow$ & CLIP Score $\uparrow$ & User Study (\%)   \\
    \midrule
    Depth Opt. & 0.6834 & 0.4180  & 0.2647 &  20.8 \\
    GenZI      & 1.3500 & 0.3542  & 0.2633 &  18.4 \\
    \textbf{Ours}       & \textbf{0.1755} & \textbf{0.0689}  & \textbf{0.2695} & \textbf{60.9} \\
    \bottomrule
  \end{tabular}
\end{table}

\begin{table}
  \caption{Quantitative comparison of text-guided multi-human generation. Our model consistently outperforms baseline methods across all evaluation metrics, demonstrating robust performance.}
  \label{multi_people_table}
  \centering
  \begin{tabular}{lcccc}
    \toprule
    &  & \multicolumn{3}{c}{Success Rate (\%)} \\
    \cmidrule(lr){3-5}
    Method     & CLIP Score $\uparrow$  & 3 human & 4 human & 5 human   \\
    \midrule
    Depth Opt. & 0.2510 & 33.3 & 18 & 13\\
    GenZI      & 0.2584 & 67.3 & 41.4 & 22 \\
    \textbf{Ours}       & \textbf{0.2685} &\textbf{100.0} & \textbf{100.0} & \textbf{100.0} \\
    \bottomrule
  \end{tabular}
\end{table}

\subsection{Results}
\label{sec:result}
Qualitative results on our model output, along with baseline outputs is shown in ~\figref{fig:qual_two_person}. Since both baseline models leverage the knowledge of 2D image diffusion model to generate plausible HOIs, their performance depends heavily on the inpainting output. In scenes where the object is actively used and in direct contact with a human, inpainting quality degrades; the target object is frequently missing from the output, which in turn leads to poor optimization quality. Additionally, as the number of people increases, the percentage of inpainting where it fails to generate the whole $N$ number of people also increases. The multi-view consistency of each human also further deteriorates as the number of people increases.

Aside from inpainting quality, both models suffer from lack of 3D HOI and HHI knowledge. Depth Opt. generates implausible human position relative to the object, due to depth estimation error. The multi-view SMPL-X parameter optimization in GenZI is sensitive to consistent body pose in each view, and can lead to poor body pose output quality. On the contrary, our generation results show high quality HHOIs in various object categories in both static and dynamic scenes.

~\tabref{two_people_table} shows quantitative results evaluating the realism on generating two people in action with object. Compared to baseline models, our model achieves significantly higher score in body pose and distance FD, implying our model can produce more realistic and natural HHOIs, that resemble those in real world environment. Note that due to background scene rendering, images with different HHOI generation output may have similar CLIP embeddings, which may be the cause of all 3 models producing CLIP scores in close range. This suggests FD of body pose and interpersonal distance is a more reliable method in evaluating how realistic each output is. Nevertheless, our model beats the baseline model in CLIP score by a slight margin. Our model also outperforms baseline models in multi-human generation. ~\tabref{multi_people_table} shows that our model achieves higher scores in CLIP score and success rate when generating more than 3 people, implying a more realistic and robust generation.

We show the physical plausibility metrics of our model output against the baselines in ~\tabref{physics_quant}. Our model outperforms baseline models in contact distance metric, implying our model output achieves more precise contact with the object compared to baseline.
For the human-human and human-object penetration ratios, our model demonstrates superior performance in dyadic human generation, whereas Depth Opt. achieves comparable results when generating three or more people. However, this apparent performance stems from Depth Opt.'s tendency to place humans at excessively large distances from objects, a consequence of the instability in depth estimation. When combined with Human Mesh Recovery, this results in low penetration metrics but unrealistic spatial arrangements in HHOI scenarios. The poor contact distance scores of Depth Opt. further substantiate this observation.

\subsection{Applications}

\begin{figure}[htbp]
    \centering
    \includegraphics[width=\linewidth, trim={0cm 0.0cm 0cm 0cm},clip]{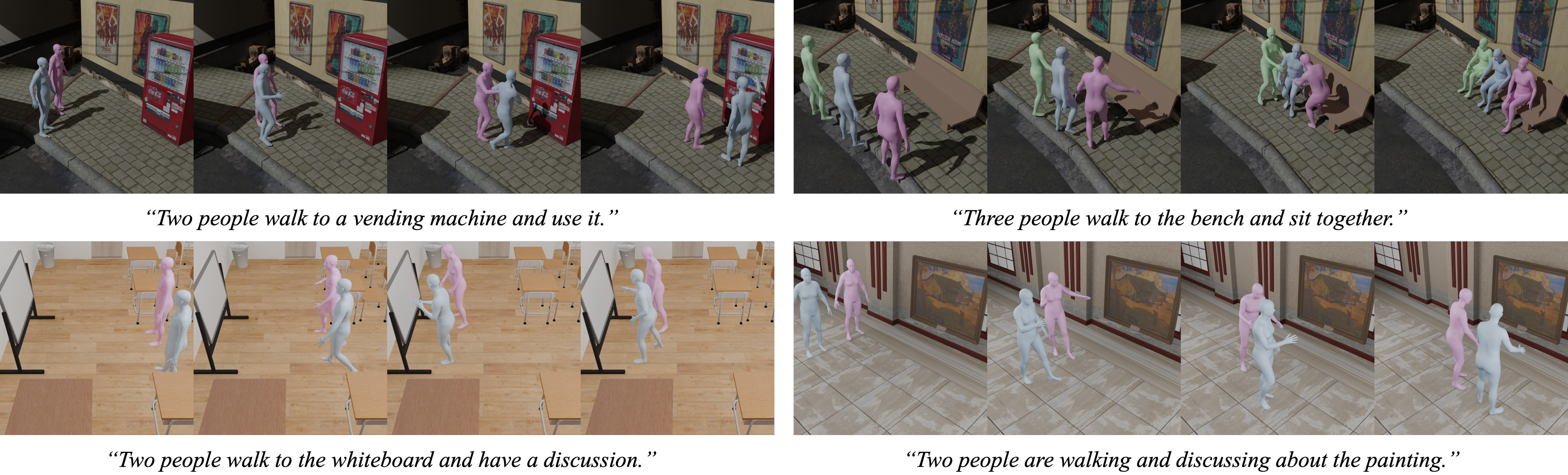}
    \caption{Motion in-betweening outputs from DNO and InterGen, given a naive standing pose as the start frame constraint and our HHOI generation output as the end frame constraint.
    }
    \label{fig:app}
\end{figure}

\begin{table}[t]
\centering
\caption{Quantitative comparison on physical plausibility metrics of our model and baseline outputs.}
\label{physics_quant}
\small
\begin{tabular}{cl|cccc}
\toprule
Metric & Method & 2 Human &3 Human & 4 Human & 5 Human \\
\midrule
\multirow{3}{*}{\parbox{2.5cm}{\centering Human-Human\\ Penetration Ratio $\downarrow$\\($\times$1000)}} 
& Depth Opt. & 13.91 & \textbf{14.06} & \textbf{19.35} & 26.16 \\
& GenZI & 13.91 & 24.60 & 21.55 & 49.06 \\
& \textbf{Ours} & \textbf{7.90} & 15.68 & 19.82 & \textbf{25.69} \\
\midrule
\multirow{3}{*}{\parbox{2.5cm}{\centering Human-Object\\ Penetration Ratio $\downarrow$\\($\times$1000)}}
& Depth Opt. & 11.48 & \textbf{4.15} & 9.41 & \textbf{1.20} \\
& GenZI & 60.71 & 15.64 & 11.93 & 8.91 \\
& \textbf{Ours} & \textbf{5.49} & 7.74 & \textbf{7.19} & 10.18 \\
\midrule
\multirow{3}{*}{\parbox{2.5cm}{\centering Contact Distance $\downarrow$\\ (m)}}
& Depth Opt. & 0.666 & 1.992 & 4.415 & 2.736 \\
& GenZI & 0.103 & 0.126 & 0.537 & 0.389 \\
& \textbf{Ours} & \textbf{0.029} & \textbf{0.031} & \textbf{0.028} & \textbf{0.026} \\
\bottomrule
\end{tabular}
\end{table}

The capability of our model to generate plausible human-human-object interactions (HHOIs) involving multiple individuals from textual prompts provides a foundation for a range of future research directions. One promising avenue is generating multiple human motion in the presence of objects in the scene. In this way, we leverage our sampled human configurations as conditioning inputs for motion-in-betweening tasks. 

Diffusion-Noise-Optimization (DNO)~\cite{karunratanakul2023dno} performs various motion-related editing tasks, using existing motion-diffusion models as a motion prior. The output of naive motion generation is compared with the condition to calculate joint loss. By changing the sampling process to ODE, we can obtain a motion sample from the latent noise deterministically and retrieve the latent noise from the motion output. Since this process is deterministic, we can backpropagate the joint loss to the inversion process and optimize the latent noise to minimize the joint loss. 

Human samples generated by our model are utilized as conditional inputs to guide the optimization of the motion diffusion model InterGen~\cite{liang2024intergen}. By integrating object interaction cues from our sampled HHOIs with motion priors from diffusion-based models, we enable the synthesis of natural multi-human motions that exhibit plausible interactions with objects in the environment. As illustrated in Fig.~\ref{fig:app}, our method allows precise positioning of generated humans and object-specific poses that are not observed in existing models.

\section{Discussion}
\label{sec:conclusion}

In this paper, we propose a method to model Human-Human-Object Interactions (HHOIs) using score-based generative models. By combining separately trained HOI and HHI models, we introduce a novel sampling strategy that enables the generation of an arbitrary number of people interacting with an object.
To train and evaluate our model, we construct a new HHOI dataset by capturing additional samples, together with synthetic data generated by our pipeline. We demonstrate that our method can synthesize realistic multi-human interaction with diverse objects.
As shown in our applications, this capability enables downstream tasks such as interaction-aware motion generation and provides a foundation for future research in multi-agent embodied intelligence.

As a limitation, our score-based model cannot directly learn HHOIs from datasets dedicated solely to HOI or HHI, due to distributional discrepancies in human configurations. Extending our framework to effectively leverage such datasets remains an open research question.

\section*{Acknowledgements}

This work was supported by NRF grant funded by the Korean government (MSIT) [No. RS-2022-NR070498 and  RS-2025-25396144], and IITP grant funded by the Korea government (MSIT) [No. RS-2024-00439854, No. RS-2025-25441838, No. RS-2021-II211343,  No. RS-2025-25442338, and No.2022-0-00156]. H. Joo is the corresponding author.

\small  
\bibliographystyle{abbrvnat}
\bibliography{main}

\newpage
\appendix

\begin{center}
  {\LARGE \textbf{Appendix}\par}
\end{center}
\vspace{1em}

\section{Data Collection Details}
\label{sec:data_collection}

\subsection{Multiview HHOIs Data Capture System}
\label{sec:capture}

\begin{figure}[h]
    \centering
    \includegraphics[width=\linewidth, trim={0 0 0 0},clip]{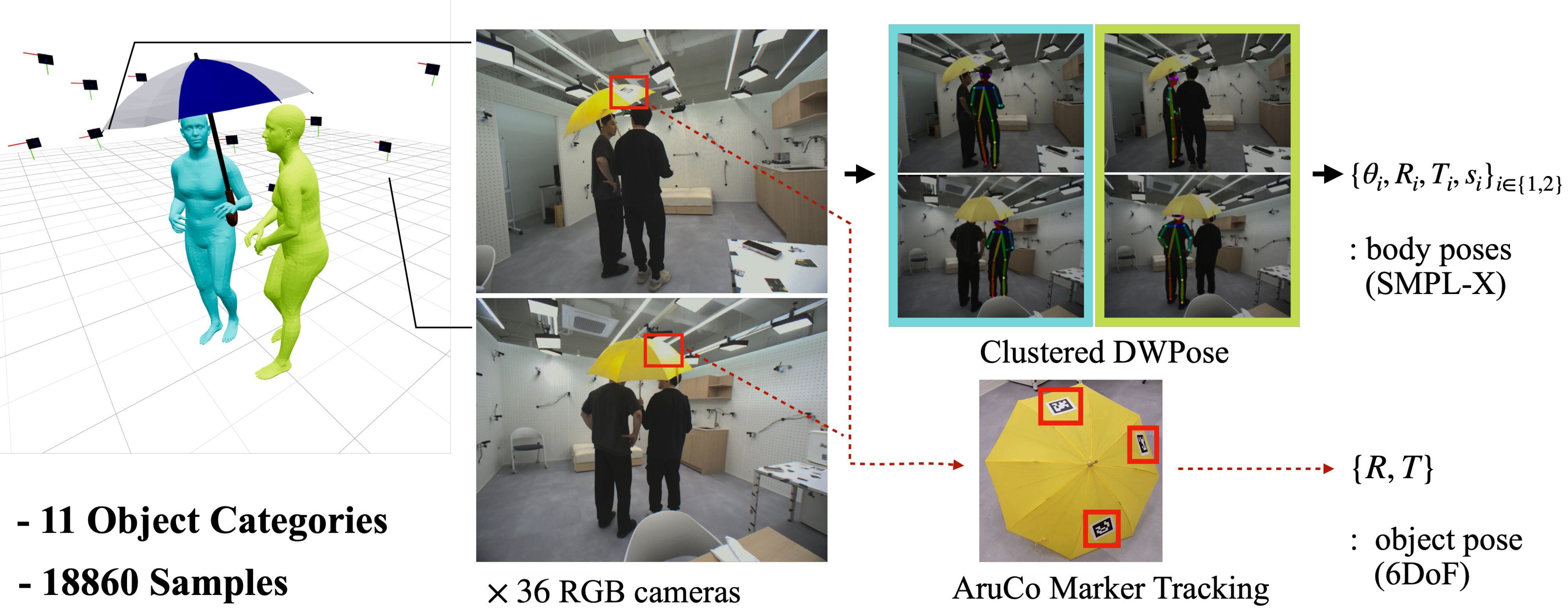}
    \caption{\textbf{HHOIs Capture System Overview.} We capture Human-Human-Object Interactions (HHOIs) with our multiple camera capture system. The object and human poses are tracked with AruCo markers~\cite{garrido2014automatic_aruco} and DWPose~\cite{dwpose} respectively.}
    \label{fig:capture_sys}
\end{figure}

To capture Human-Human-Object Interactions (HHOIs), we adopt a multi-camera setup inspired by Panoptic Studio~\cite{joo2015panoptic}. As shown in ~\figref{fig:capture_sys}, our system is composed of 36 synchronized RGB cameras to ensure accurate human pose estimation, even under severe occlusions during interaction. For tracking the 6-DoF pose of the interacting object, we attach ArUco markers~\cite{garrido2014automatic_aruco} on the object surface, which provide robust and efficient object pose annotation throughout the interaction sequence. 

We use DWPose~\cite{dwpose} to extract 2D human keypoints from each camera view. To associate human detections across views, we incrementally cluster the detected people in 3D by minimizing the overall triangulation error, ensuring consistent identity assignment. Once the keypoints are clustered, we follow the preprocessing pipeline of PaMIR~\cite{mvsmplifyx} to fit SMPL-X parameters for each individual. This involves optimizing body pose, scale, translation, and rotation with regularization from learned priors~\cite{smplx}, resulting in plausible 3D human poses.

For object tracking, we begin by placing the target object at the center of the capture system to align its template mesh within the coordinate system of our multi-camera setup. 
After capture, we reconstruct a 3D Gaussian Splatting (3D-GS)~\cite{kerbl3Dgaussians} scene from the initial frames and manually align the template mesh to the physical object in the scene to determine its object pose. For dynamic objects, subsequent poses are computed by applying 6-DoF transformations from the ArUco markers.

During the data capture process, two actors are provided with high-level instructions for the interaction scenario while maintaining flexibility in execution to encourage natural behaviors and a diverse range of human poses. After recording, we post-process the captured sequences by categorizing them into fine-grained sub-scenarios based on interaction type and body configuration. To ensure data quality, we filter out frames where SMPL-X fitting is unreliable due to identity clustering failure, occlusion, or low keypoint confidence, resulting in a clean and robust dataset for downstream tasks.

\subsection{Data Preprocessing}
\label{sec:data}

For the collected HHOI data from CORE4D~\cite{liu2024core4d} and our multiview capture system, each sample includes the object instance, as well as the SMPL-X translation, rotation, and body pose parameters for the two interacting individuals. To prepare the data for our model, we first normalize the object's position and orientation to the origin and a canonical frame. The human poses are then transformed accordingly, such that the resulting SMPL-X parameters represent each individual's body pose, translation, and rotation relative to the canonical object frame. These parameters are used as input to our HOI model.

For the HHI model, we normalize one individual’s position to the origin and its rotation to the identity (i.e., zero in the SMPL-X global rotation parameter), effectively expressing the second human's translation and rotation relative to the first. We repeat this process by switching the reference human, resulting in two HHI data samples per HHOI frame. After preprocessing, the data is split into training and test sets with a 9:1 ratio. This test set serves as the ground-truth distribution for computing the Fréchet Distance in Sec.~\ref{sec:experiments}, enabling quantitative comparison of the realism of our generated results against the baseline methods.

For synthetic data, ComA directly generates HOI data in a format compatible with our model. Synthetic HHI samples are derived from the Human Mesh Recovery output with the above explained normalization procedure.

\label{sec:stat}
\begin{figure}[t]
    \centering
    \includegraphics[width=\linewidth, trim={0 0 0 0},clip]{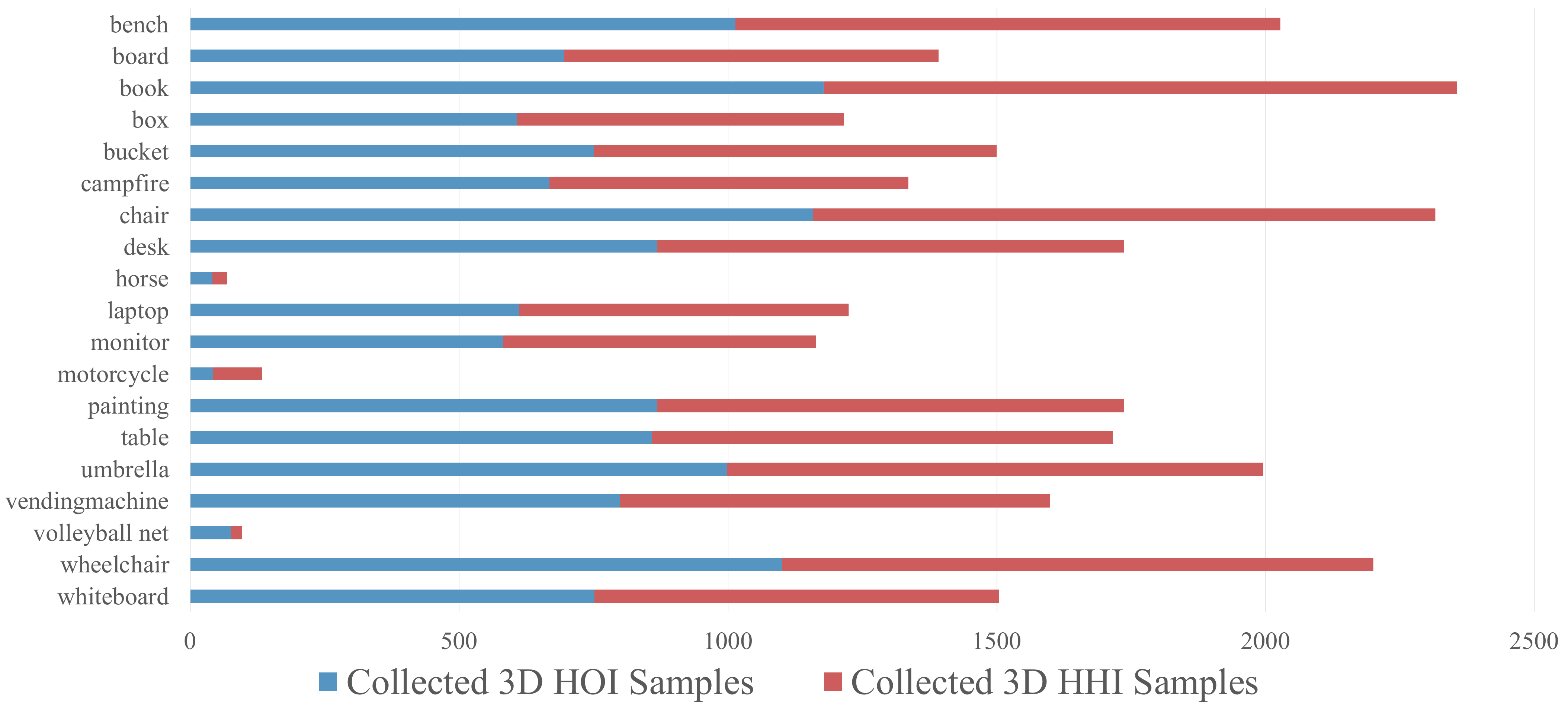}
    \caption{\textbf{Statistics on Collected Data Samples.} Our dataset is constructed by integrating data from CORE4D and our multiview capture system, alongside synthetic samples generated via our data generation pipeline.}
    \label{fig:data_stat}
\end{figure}

\subsection{Data Statistics}

In total, we collect 13,669 3D HOI samples and 13,650 3D HHI samples, spanning 19 object categories. Of these, 5 categories—board, box, bucket, chair, and desk—are obtained from the CORE-4D dataset. 11 categories—bench, book, campfire, laptop, monitor, painting, table, umbrella, vending machine, wheelchair, whiteboard—are captured using our multi-view capture system. The remaining 3 categories—horse, motorcycle, and volleyball net—are generated using our synthetic data pipeline. 27,020 samples are collected from CORE4D and capture system, and 299 samples are generated from synthetic pipeline. Detailed statistics for each category are shown in Fig.~\ref{fig:data_stat}.

\section{Implementation Details}
\label{sec:impl_details}

\subsection{HHOI Diffusion Architecture}
\label{sec:arch}

\begin{figure}[t]
    \centering
    \includegraphics[width=\linewidth, trim={0 0 0 0},clip]{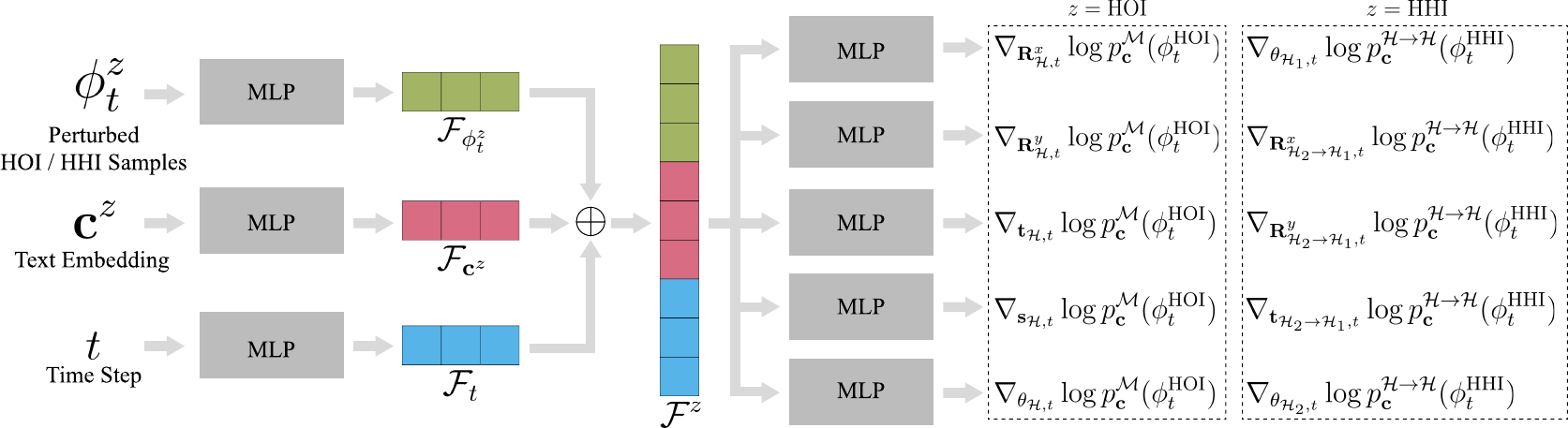}
    \caption{\textbf{
    HHOI Diffusion Architecture.} HHOI diffusion consists of two disjoint diffusion models: HOI diffusion and HHI diffusion. Although the overall structure of each diffusion model is the same, they are implemented with separate networks due to their different target distributions. Each network learns the score function of the HOI or HHI distribution, respectively.
    }
    \label{fig:impl_detail}
\end{figure}

We adopt the MLP-based score network from GenPose~\cite{genpose}, which is originally designed to predict 6D poses of point clouds, as the backbone of our HHOI diffusion model. We once again emphasize that HHOI diffusion is composed of two disjoint diffusion models: HOI diffusion and HHI diffusion. These two diffusion models do not share parameters and are trained separately. However, as shown in Fig.~\ref{fig:impl_detail}, the overall architecture of the two diffusion models is the same. First, separate MLPs are used as feature extractors for the HOI or HHI sample, the CLIP text embedding, and the time step, respectively. The resulting features are denoted as $\mathcal{F}_{\phi_t^z} \in \mathbb{R}^{256}$ for the HOI or HHI sample, $\mathcal{F}_{\mathbf{c}^z} \in \mathbb{R}^{128}$ for the CLIP text embedding, and $\mathcal{F}_{t} \in \mathbb{R}^{128}$ for the time step. The features are concatenated into a single vector $\mathcal{F}^z \in \mathbb{R}^{512}$, which is then passed to other MLPs to predict the score functions. Each diffusion model employs distinct MLPs to predict the score functions corresponding to each component in $\phi^z_t$. In particular, following GenPose, 6D rotations are decomposed into two 3D vectors, and separate score functions are learned for each. The total number of trainable parameters in the HHOI diffusion model is 10.2M, with approximately 5.1M in each of the HOI and HHI diffusion models.

\begin{figure}[t]
    \centering
    \includegraphics[width=\linewidth, trim={0 0 0 0},clip]{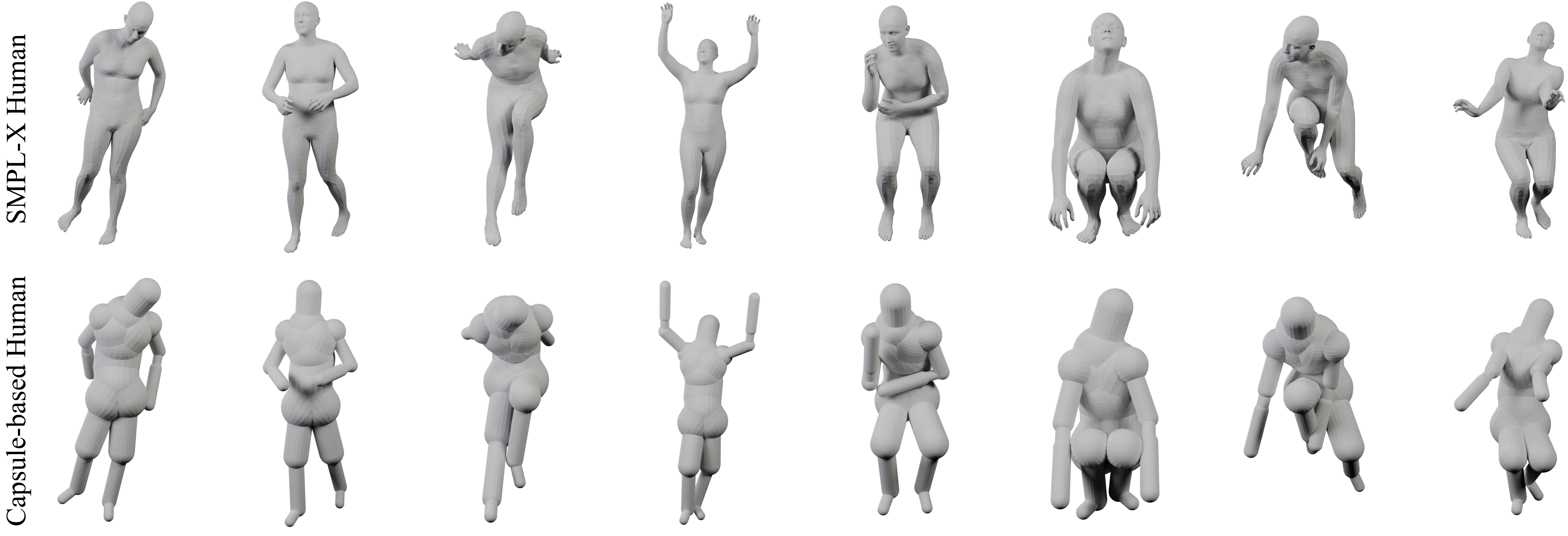}
    \caption{\textbf{Capsule-based Approximation of SMPL-X Human.} Our capsule-based human approximation models SMPL-X humans well, enabling a computationally cheap collision loss formulation.}
    \label{fig:capsule}
\end{figure}

\subsection{Capsule-based Human for Collision Loss}
\label{sec:capsule_appendix}

In Sec.~\ref{sec:hhoi_infer}, we approximate each human with a 24-capsule proxy to define a computationally efficient collision loss. A capsule is a set of points that are equidistant from a line segment called an axis segment. Consequently, collisions between two capsules are determined simply by checking whether the minimum distance between their axis segments is smaller than the sum of their radii.

Given a body pose embedding $\theta_\mathcal{H}$, we construct the capsule proxy through the following steps: (1) Decode $\theta_\mathcal{H}$ with a body pose decoder to obtain the actual 21$\times$6D body pose, and apply forward kinematics to obtain the positions of the 22 human joints including root; (2) Consider the 21 bones that connect each joint to its parent joint as the axis segment of each capsule; (3) Define 3 additional axis segments for the capsules corresponding to the hands and head, utilizing the two wrists and head joints; (4) Apply pre-learned per-capsule radii to obtain the 24-capsule proxy. Fig.~\ref{fig:capsule} shows that our capsule-based human approximation works well.

To obtain per-capsule radii, we reuse the 922K human body pose samples used to train the body pose encoder-decoder described in Sec.~\ref{sec:hhoi_formulation}. For each pose sample, we generate the corresponding SMPL-X mesh and its capsule-based approximation, uniformly sample 5000 surface points from each, and then optimize a set of per-capsule radii by minimizing the Chamfer distance between the two point sets.

\subsection{Hyper Parameter Settings}
\label{sec:hp}

We train both the HOI and HHI diffusion models for 20,000 epochs with a batch size of 500. The learning rate is initialized at 1e-2 and decays by a factor of 0.999 at each step, with a lower bound of 7e-4.
For guided HHOI sampling, we set $\lambda_1 = \min(100000, \frac{100}{t^2})$ and $\lambda_2 = \min(1600000, \frac{1600}{t^2})$, which correspond to the weight terms defined in \eqnref{eqn:reverse_ode_full}. To ensure that the inconsistency loss and collision loss have meaningful effects, we apply these losses starting from the time step $t = 0.5$. Consequently, HHOI samples are obtained by solving the following PF ODE:
\begin{equation}\label{eqn:reverse_ode_full_detail}
    \begin{gathered}
       \phi_{1}^{z, i} \sim \mathcal{N}(\mathbf{0}, \sigma_{\text{max}}^2\mathbf{I}_z), \\
       \frac{d\phi_{t}^{z, i}}{dt} = -\sigma(t)\dot\sigma(t)\Psi_{\Theta^{z}}(\phi_{t}^{z, i}, t | \mathbf{c}^z, z), \;\;  t \in (0.5, 1.0], \\
       \frac{d\phi_{t}^{z, i}}{dt} = -\sigma(t)\dot\sigma(t)\Psi_{\Theta^{z}}(\phi_{t}^{z, i}, t | \mathbf{c}^z, z) + 
       \lambda_1\nabla_ {\phi_{t}^{z, i}}\mathcal{L}_{\text{inc}}(\Phi_t) +
       \lambda_2\nabla_{\phi_{t}^{z, i}}\mathcal{L}_{\text{col}}(\Phi_t), \;\;  t \in [\epsilon, 0.5].
    \end{gathered}
\end{equation}

\subsection{Text Augmentation for Training}
\label{sec:text_aug}

\begin{figure}[t]
    \centering
    \includegraphics[width=\linewidth, trim={0 0 0 0},clip]{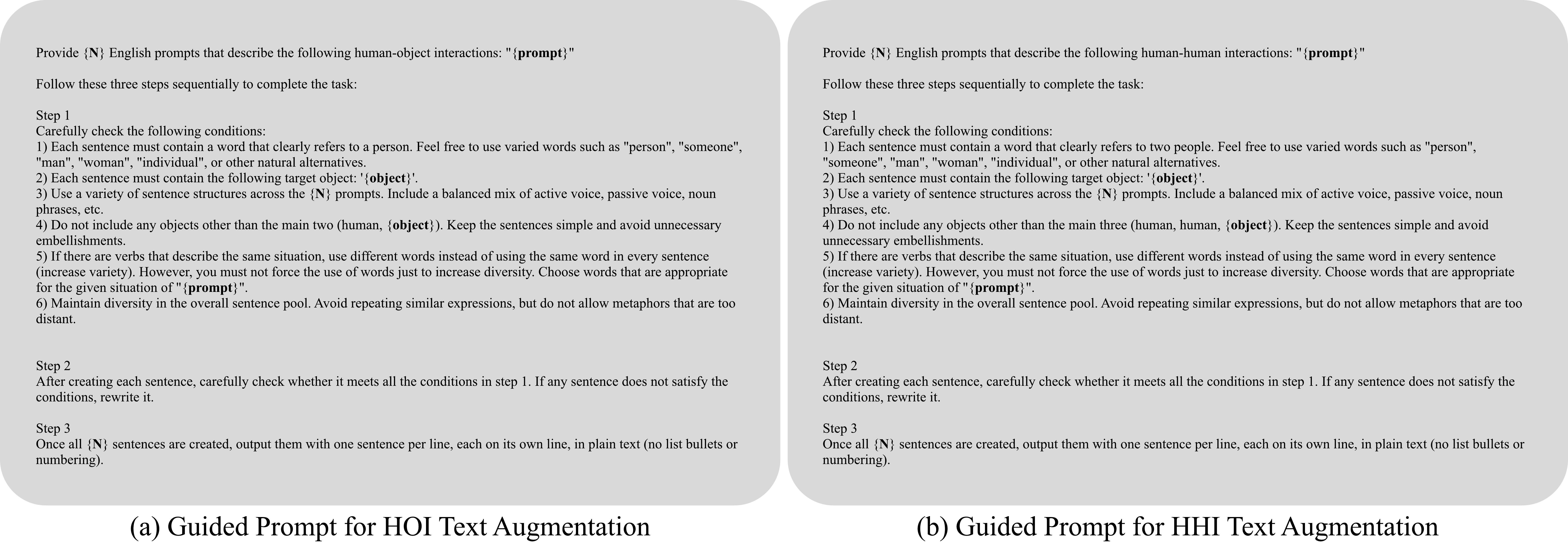}
    \caption{\textbf{
    Guided Prompt Provided to LLM for Text Augmentation.} (a) Guided prompt for HOI text prompt augmentation. (b) Guided prompt for HHI text prompt augmentation.
    }
    \label{fig:guided_text}
\end{figure}

We adopt the LLM-based text augmentation method proposed in ~\cite{oor} to train our text-conditioned HHOI diffusion model. Fig.~\ref{fig:guided_text} shows the detailed guided prompts to be provided to LLM~\cite{openai2022chatgpt}. The guided prompt instructs the LLM to produce diverse sentence structures, varied verbs, and concise sentences without flowery language. 
We demonstrate in Sec.~\ref{sec:additional_results} that our model generalizes to unseen text.

\section{Experiments Details}
\label{sec:exp_detail}

\subsection{Baseline Methods}
\label{sec:baseline}

\noindent \textbf{GenZI} We modify GenZI ~\cite{genzi} method to generate multiple number of humans in the given scene, instead of one. We follow GenZI's pipeline in selecting the views to render and inpaint the scene. In the inpainting process, we change the input text prompt to the inpainting pipeline to suit our desired number of humans, and correpondingly set the tokens used in dynamic masking. AlphaPose~\cite{fang2022alphapose} detects the 2D joint positions of generated humans in each image. Given that each image contains multiple individuals, establishing cross-view correspondence for each human is needed to reconstruct 3D humans from the multi-view images. We estimate the correspondence by evaluating possible correspondence combinations and selecting the set that minimizes the reprojection error. Subsequently, the original objective function of GenZI is calculated independently for each individual, and the final loss is obtained by summing these individual losses. For the iterative refinement, we load all the generated humans and render the silhouette to use as the inpainting mask.

\noindent \textbf{Depth Opt.} For a simple but effective baseline, we employ Human Mesh Recovery and Depth Estimation method to generate 3D humans from text prompt and scene. Following the GenZI framework, we render a single-view image of the scene and inpaint multiple individuals based on a given text prompt. Then, we reconstruct the 3D human poses from the generated image using multi-HMR~\cite{multihmr}. With the generated SMPL-X humans and the scene, we render to obtain the image and depth. Then, we use Depth-Pro~\cite{ml-depth-pro} to estimate the depth from the image. The image is segmented into scene-only and human-only pixels. For the scene-only regions, we compute the difference between the estimated depth and the ground-truth depth obtained from the renderer. This difference is averaged across all scene-only pixels and used as an offset, which is then applied to the estimated depths in the human-only regions. Finally, we compute the ratio between the offset-corrected estimated depth and the rendered depth for the human-only pixels, and apply this ratio to scale the generated 3D human reconstructions accordingly. 

\subsection{User Study}
\label{sec:user_study}
\begin{figure}[t]
    \centering
    \includegraphics[trim=0 485 350 0, clip, width=1.0\linewidth]{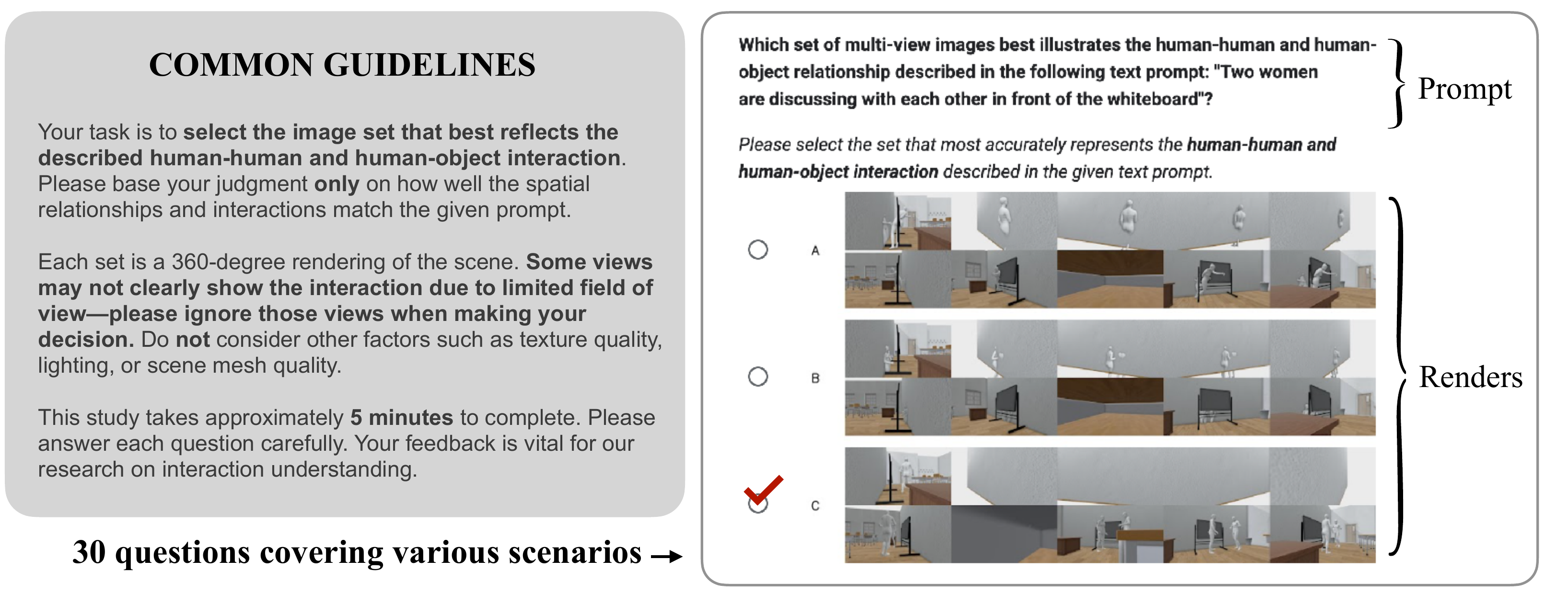}
    \caption{\textbf{Questionnaire for User Study.} Participants select the multi-view image that best depicts the human-human and human-object relationship.
    }
    \label{fig:user_study}
\end{figure}

We conduct a user study to evaluate how effectively our model captures human-human and human-object interactions. We collect responses from 97 participants via CloudResearch~\cite{cloudresearch}, offering a reward credit of \$1.50 per participant. The study comprises 30 questions covering 10 objects, each associated with three randomly ordered interaction scenarios—one from our model and two from baseline methods. For each question, participants are presented with 10 multi-view renderings of the HHOI scene, showing the humans, object, and environment as generated by each method. They are asked to select the rendering that best illustrates the given text prompt. The structure of the questionnaire is shown in Fig.~\ref{fig:user_study}.

\section{Additional Qualitative Results}
\label{sec:additional_results}

\begin{figure}[t]
    \centering
    \includegraphics[width=0.92\linewidth, trim={0 0 0 0},clip]{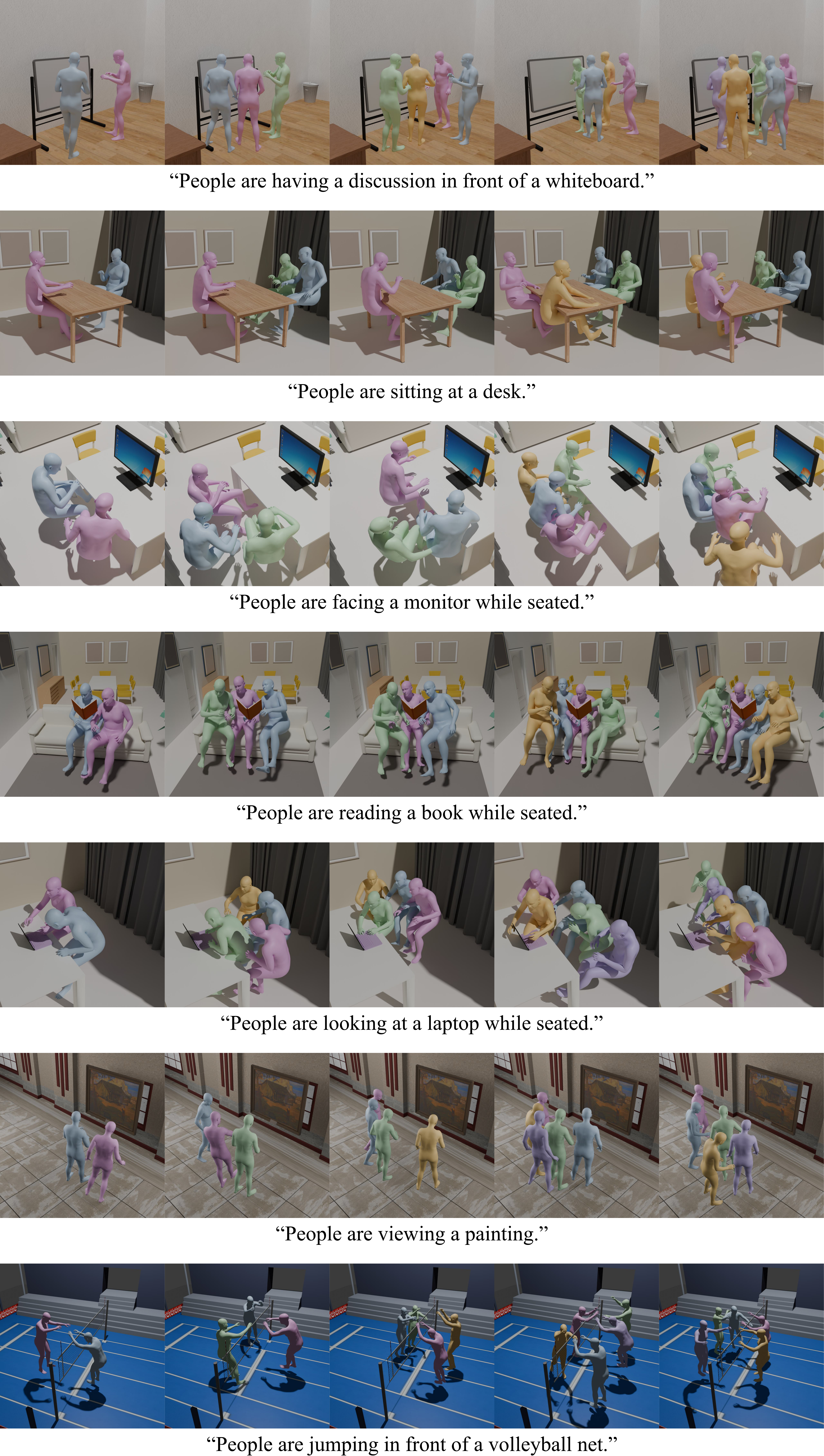}
    \caption{\textbf{
    Additional HHOI Qualitative Results.} 
    }
    \label{fig:add_qual}
\end{figure}

Fig.~\ref{fig:add_qual} shows the HHOI results obtained using the advanced sampling introduced in Sec.~\ref{sec:hhoi_infer}. Our method performs well even in multi-human cases involving more than two individuals across diverse categories.

\begin{figure}[t]
    \centering
    \includegraphics[width=\linewidth, trim={0 0 0 0},clip]{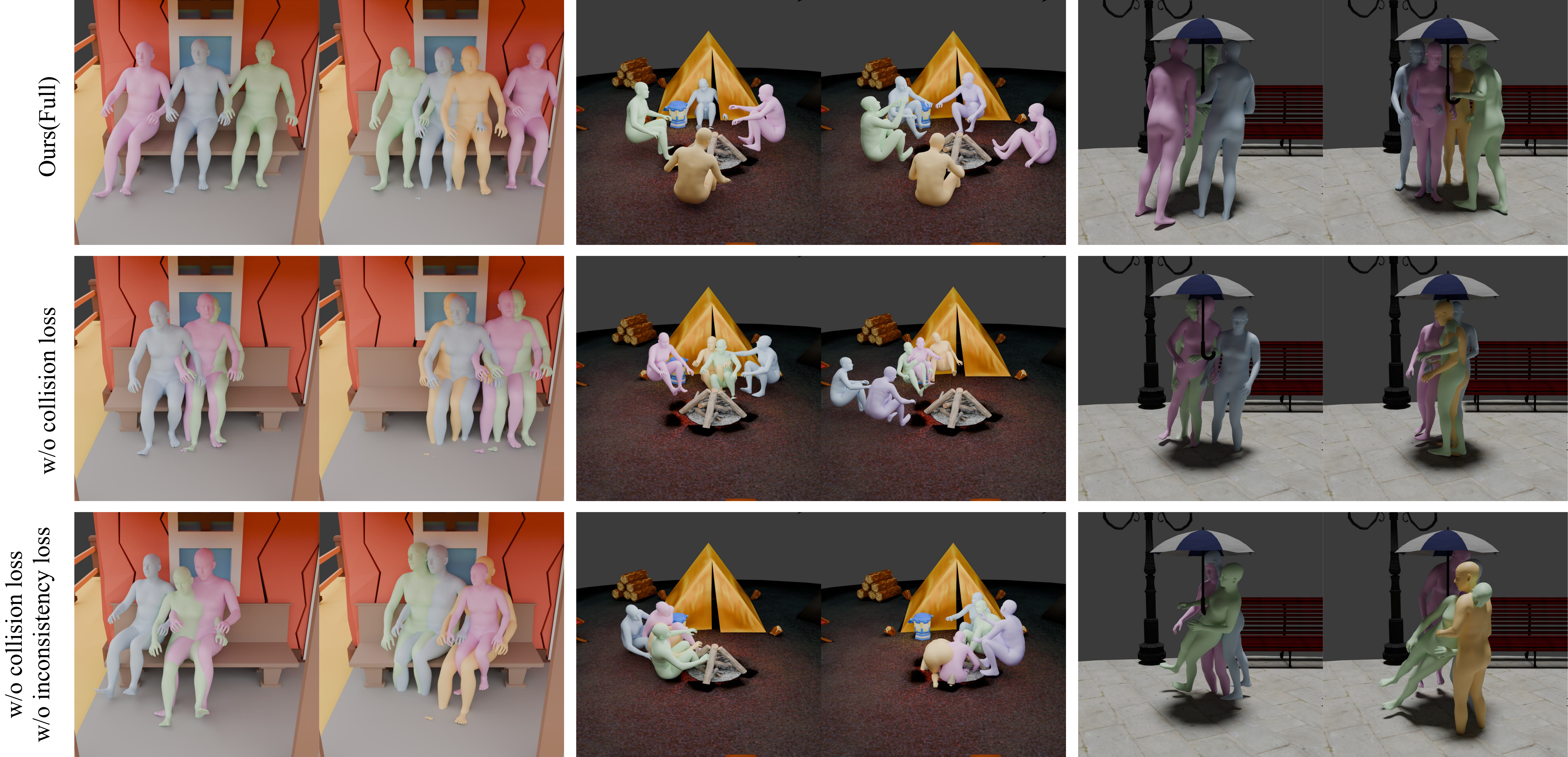}
    \caption{\textbf{
    Ablation Study for Guided HHOI Sampling.} 
    }
    \label{fig:ablation}
\end{figure}

We additionally provide ablation results for our guided HHOI sampling. Fig.~\ref{fig:ablation} presents the HHOI sampling results for three versions: our full method, a version without collision loss, and a version without both collision and inconsistency losses. Without the collision loss, in cases involving more than three individuals, collisions are more likely to occur between humans whose HHI relationships are not explicitly specified. 
Removing the inconsistency loss disrupts the integration of HOI and HHI, producing less plausible HHOI results.

\begin{figure}[t]
    \centering
    \includegraphics[width=\linewidth, trim={0 0 0 0},clip]{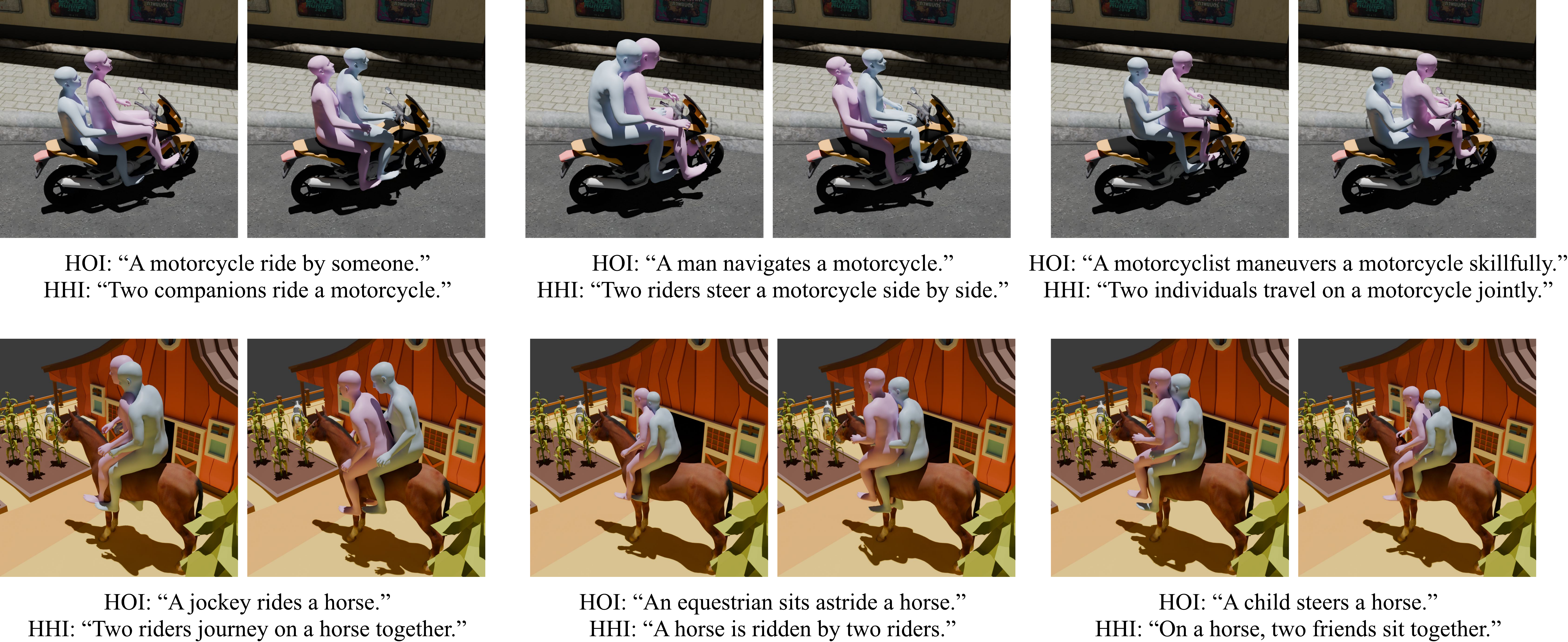}
    \caption{\textbf{
    Qualitative Results of HHOI Sampling for Unseen Text Prompts} 
    }
    \label{fig:unseen_text}
\end{figure}

Finally, to demonstrate the effectiveness of our text augmentation, we present in Fig.~\ref{fig:unseen_text} the HHOI sampling results for unseen text prompts.

\section{Licenses and Rights}

The following 3D models were used in this work for dataset and rendering. All assets are publicly available under the Creative Commons Attribution (CC-BY 4.0 or CC BY-NC 4.0) license or cited from academic datasets. Attribution is provided below where required.

\begin{itemize}
  \item \textbf{Bench} \\
  \textit{Source}: \url{https://fab.com/s/5cfe864b44e1} \\
  \textit{License}: CC BY 4.0 \\
  \textit{Attribution}: "Bench" is licensed under CC BY 4.0.

  \item \textbf{Book} \\
  \textit{Title}: Book opened \\
  \textit{Source}: \url{https://skfb.ly/6s9uU} \\
  \textit{License}: CC BY 4.0 \\
  \textit{Attribution}: "Book opened" by Jiří Kuba is licensed under CC BY 4.0.

  \item \textbf{Campfire} \\
  \textit{Title}: Campfire Wood Survival Warm and Light \\
  \textit{Source}: \url{https://skfb.ly/6QYoY} \\
  \textit{License}: CC BY 4.0 \\
  \textit{Attribution}: "Campfire Wood Survival Warm and Light" by digrafstudio is licensed under CC BY 4.0.

  \item \textbf{Horse} \\
  \textit{Title}: Horse basemesh \\
  \textit{Source}: \url{https://skfb.ly/oZoW8} \\
  \textit{License}: CC BY 4.0 \\
  \textit{Attribution}: "Horse basemesh" by Sid Ahearne is licensed under CC BY 4.0

  \item \textbf{Laptop} \\
  \textit{Dataset}: SAPIEN (Object ID: 10211) ~\cite{xiang2020sapien} 
  
  \item \textbf{Monitor} \\
  \textit{Title}: Desktop Computer \\
  \textit{Source}: \url{https://skfb.ly/6RCNx} \\
  \textit{License}: CC BY 4.0 \\
  \textit{Attribution}: "Desktop Computer" by Tytan is licensed under CC BY 4.0.

  \item \textbf{Motorcycle} \\
  \textit{Title}: Motorcycle \\
  \textit{Source}: \url{https://skfb.ly/6oEPN} \\
  \textit{License}: CC BY 4.0 \\
  \textit{Attribution}: "Motorcycle" by Silas6 is licensed under CC BY 4.0

  \item \textbf{Painting} \\
  \textit{Title}: Birthplace Painting \\
  \textit{Source}: \url{https://skfb.ly/6zNAL} \\
  \textit{License}: CC BY 4.0 \\
  \textit{Attribution}: "Birthplace Painting" by Znyth Technologies is licensed under CC BY 4.0.

  \item \textbf{Table} \\
  \textit{Title}: Kitchen table \\
  \textit{Source}: \url{https://skfb.ly/6Goxs} \\
  \textit{License}: CC BY 4.0 \\
  \textit{Attribution}: This work is based on "Kitchen table" by tahax licensed under CC BY 4.0.

  \item \textbf{Umbrella} \\
  \textit{Title}: Umbrella \\
  \textit{Source}: \url{https://skfb.ly/6YpNI} \\
  \textit{License}: CC BY 4.0 \\
  \textit{Attribution}: "Umbrella" by Diccbudd is licensed under CC BY 4.0.

  \item \textbf{Vending Machine} \\
  \textit{Title}: Vending Machine \\
  \textit{Source}: \url{https://skfb.ly/6ZtVQ} \\
  \textit{License}: CC BY 4.0 \\
  \textit{Attribution}: "Vending Machine" by yashwanthantony9542 is licensed under CC BY 4.0

  \item \textbf{Volleyball Net} \\
  \textit{Title}: volleyball net \\
  \textit{Source}: \url{https://skfb.ly/oYMHq} \\
  \textit{License}: CC BY 4.0 \\
  \textit{Attribution}: "volleyball net" by otyken is licensed under CC BY 4.0

  \item \textbf{Wheelchair} \\
  \textit{Title}: Wheelchair \\
  \textit{Source}: \url{https://skfb.ly/ouV8F} \\
  \textit{License}: CC BY 4.0 \\
  \textit{Attribution}: "Wheelchair" by Fine\_poultry is licensed under CC BY 4.0.

  \item \textbf{Whiteboard} \\
  \textit{Source}: \url{https://www.fab.com/listings/398f36d0-bd12-4d93-b348-0f02f1677eae} \\
  \textit{License}: CC BY 4.0 \\
  \textit{Attribution}: "Whiteboard" is licensed under CC BY 4.0.

\end{itemize}

The following 3D models were used in this work as a background mesh to render our generation result, and for the baseline model implementation. Attribution is provided below where required.

\begin{itemize}
    \item \textbf{Low Poly Farm V2} \\
    \textit{Source}: \url{https://skfb.ly/6QYJI} \\
    \textit{License}: CC BY 4.0 \\
    \textit{Attribution}: "Low Poly Farm V2" by EdwiixGG is licensed under CC BY 4.0

    \item \textbf{Bangkok City Scene} \\
    \textit{Source}: \url{https://skfb.ly/6GxUO} \\
    \textit{License}: CC BY 4.0 \\
    \textit{Attribution}: "Bangkok City Scene" by ArneDC is licensed under CC BY 4.0

    \item \textbf{Living Room} \\
    \textit{Source}: \url{https://skfb.ly/6wYHE} \\
    \textit{License}: CC BY 4.0 \\
    \textit{Attribution}: "Living Room" by Taranpreet is licensed under CC BY 4.0

    \item \textbf{Camp Scene, Free Download} \\
    \textit{Source}: \url{https://skfb.ly/6SnYV} \\
    \textit{License}: CC BY NC-4.0 \\
    \textit{Attribution}: "Camp Scene, Free Download" by Bento is licensed under CC BY NC-4.0

    \item \textbf{Low Poly Simple Hallway Room} \\
    \textit{Source}: \url{https://skfb.ly/oyLtI} \\
    \textit{License}: CC BY 4.0 \\
    \textit{Attribution}: "Low Poly Simple Hallway Room" by jimbogies is licensed under CC BY 4.0

    \item \textbf{Livingroom} \\
    \textit{Source}: \url{https://skfb.ly/6RBx8} \\
    \textit{License}: CC BY 4.0 \\
    \textit{Attribution}: "Livingroom" by Amy is licensed under CC BY 4.0

    \item \textbf{Gallery Museum Showroom Banquet Hall} \\
    \textit{Source}: \url{https://skfb.ly/otqOY} \\
    \textit{License}: CC BY 4.0 \\
    \textit{Attribution}: "Gallery Museum Showroom Banquet Hall" by jimbogies is licensed under CC BY 4.0

    \item \textbf{Venice city scene 1DAE08 Aaron Ongena} \\
    \textit{Source}: \url{https://skfb.ly/6TptH} \\
    \textit{License}: CC BY 4.0 \\
    \textit{Attribution}: "Venice city scene 1DAE08 Aaron Ongena" by AaronOngena is licensed under CC BY 4.0

    \item \textbf{Basic Classroom} \\
    \textit{Source}: \url{https://skfb.ly/ovnsE} \\
    \textit{License}: CC BY 4.0 \\
    \textit{Attribution}: "Basic Classroom" by Kibele is licensed under CC BY 4.0

    \item \textbf{Tennis Court} \\
    \textit{Source}: \url{https://skfb.ly/6YKBs} \\
    \textit{License}: CC BY 4.0 \\
    \textit{Attribution}: "Tennis Court" by Spark Games is licensed under CC BY 4.0
    
\end{itemize}

\end{document}